\documentclass[a4paper,fleqn]{cas-sc}

\usepackage[authoryear,longnamesfirst]{natbib}
\usepackage{amsmath,amssymb,amsthm,mathtools}
\usepackage{bm}
\usepackage{graphicx}
\usepackage{booktabs}

\usepackage{placeins}
\usepackage{float}
\restylefloat*{figure}

\newtheorem{theorem}{Theorem}
\newtheorem{lemma}[theorem]{Lemma}
\newtheorem{corollary}[theorem]{Corollary}
\newtheorem{proposition}[theorem]{Proposition}

\newdefinition{definition}{Definition}
\newdefinition{remark}{Remark}

\newcommand{\dx}{\mathrm{d}x}
\newcommand{\dy}{\mathrm{d}y}
\newcommand{\R}{\mathbb{R}}

\newcommand{\Gglob}{G^{\mathrm{glob}}}
\newcommand{\Ggate}{G^{\mathrm{gate}}}

\newcommand{\diagpre}{\mathcal{P}_{\text{d}}}
\newcommand{\toeppre}{\mathcal{P}_{\text{t}}}
\newcommand{\circpre}{\mathcal{P}_{\text{c}}}

\begin{document}
\RenewDocumentCommand\printorcid{}{}
\let\WriteBookmarks\relax
\def\floatpagepagefraction{1}
\def\textpagefraction{.001}

\shorttitle{Spectral bias reduction in gated ReLU networks}
\shortauthors{V. Dolean}

\title[mode = title]{Spectral Bias Reduction in Local Gated ReLU Architectures}

\author[1]{Victorita Dolean}
\cormark[1]
\ead{v.dolean.maini@tue.nl}
\credit{Conceptualization, Methodology, Formal analysis, Software, Numerical experiments, Writing -- original draft, Writing -- review \& editing}

\affiliation[1]{organization={Department of Mathematics and Computer Science,
                              Eindhoven University of Technology},
                addressline={PO Box 513},
                city={Eindhoven},
                postcode={5600 MB},
                country={The Netherlands}}

\cortext[cor1]{Corresponding author.}

\begin{abstract}
Neural networks trained by gradient-based methods learn the low-frequency content of a
target function faster than its high-frequency content. This spectral bias is well
documented, but it is much less clear which architectural choices mitigate it and which
cannot. We study the question for shallow ReLU models linear in their trainable parameters, where
each mode is learned at a rate given by a Gram-matrix eigenvalue, and for a general
symmetric bilinear form $a(\cdot,\cdot)$, so that least-squares regression ($a=$ the $L^2$
inner product) and variational or physics-informed losses ($a=$ the $H^1$ energy form) are
two instances of one analysis.

Within this framework we propose gated local ReLU dictionaries, obtained by multiplying
each ReLU feature by a hat partition-of-unity weight, in the spirit of partition-of-unity
finite elements and overlapping Schwarz methods. The resulting Gram
matrices are block-banded for any local bilinear form, with a block-Toeplitz interior
whose matrix-valued symbol governs the spectrum. 

The two forms then behave very differently. For $L^2$ regression, the symbol is bounded
away from zero, the ratio is uniform in the number of windows, and the measured spread of
frequency-dependent learning rates falls from $1.9\times10^{5}$ to $4.5$ at matched model
size. For the $H^1$ stiffness matrix, the symbol vanishes at zero frequency: gating leaves
the $\Theta(N^2)$ conditioning intact and inverts the frequency ordering instead of
removing the bias. A preconditioner built from the same interior stencil will fix the spectral bias, showing that architecture alone is not enough.
Experiments on $1$D regression and Poisson problems confirm the dichotomy, and the
regression effect persists when the hidden-layer parameters are trained as well.
\end{abstract}



\maketitle

\section{Introduction}
\label{sec:intro}

Neural networks trained by gradient-based optimization exhibit a pronounced \emph{spectral bias}: they learn the low-frequency components of a target function substantially faster than the high-frequency ones \citep{rahaman2019spectral,xu2020frequency,basri2019convergence}. The phenomenon has been observed across architectures, datasets, and tasks, and has been linked to generalization, training efficiency, the success of neural representations, and the well-known degradation of performance of physics-informed neural networks (PINNs) on multi-scale problems \citep{tancik2020fourier,wang2021eigenvector,wang2022when}.

The dominant theoretical lens is the neural tangent kernel (NTK) \citep{jacot2018ntk}: in the infinite-width limit, training dynamics under gradient flow are governed by a fixed kernel whose eigenvalues decay rapidly with frequency, mechanically producing a low-frequency bias \citep{basri2019convergence,arora2019fine,du2018gradient,basri2020frequency}. While powerful, this viewpoint offers limited \emph{architectural} guidance: it explains why bias exists but is less direct about how to choose finite-width architectures that mitigate it. Architectural remedies in current practice -- Fourier feature mappings, sinusoidal activations, polynomial nets, wavelet/KAN networks -- are typically motivated by reshaping the NTK eigenspectrum by other means, often without finite-width spectral guarantees \citep{tancik2020fourier,sitzmann2020implicit,liu2024kan,meshir2025wavkan}.

In this work, we adopt a complementary, finite-dimensional viewpoint that puts the \emph{feature dictionary itself} at the center of the analysis. We consider shallow approximations
\begin{equation}
\label{eq:linear-in-params}
   f_\theta(x) \;=\; \sum_{\ell=1}^{N} \theta_\ell\,\varphi_\ell(x),
   \qquad \theta\in\R^N,
\end{equation}
with ReLU-based features $\{\varphi_\ell\}$, trained by gradient descent on a quadratic loss. The model is simpler than a fully trainable neural network, but it captures the main mechanism of spectral bias in a transparent form: under gradient descent the error in the eigenbasis of the Gram feature matrix $G$ decays mode-by-mode at rates dictated by its eigenvalues \citep{jacot2018ntk,mei2018meanfield,hong2022activation}. Poor spectral conditioning of $G$ therefore directly translates into a large gap between the learning rates of the low- and high-frequency modes.

Our starting point is the recent observation by \citet{hong2022activation} that, for \emph{global} ReLU feature dictionaries, the Gram matrix is closely related to a finite-element mass matrix and exhibits a quartic blow-up of its condition number with the number of features. This is precisely the source of the strong spectral bias of shallow ReLU networks. Motivated by classical ideas from numerical analysis --  in particular the partition-of-unity finite element method \citep{babuvska1997partition} and overlapping Schwarz domain decomposition \citep{Dolean:15:DDM, toselli2005domain} -- we replace the global ReLU dictionary with a \emph{gated local} variant in which each ReLU feature is multiplied by a partition-of-unity weight. The result is still linear in parameters, still expressible as a shallow ReLU-type network, but is spatially localized.

This change has a direct spectral consequence. The gated Gram matrix is \emph{block-banded}, so its spectrum can be analyzed with the theory of block-Toeplitz matrices and their matrix-valued symbols \citep{serra1999spectral,bottcher2005spectral}, a body of results that is standard in numerical linear algebra but little used in this context.

The symbol connects the architecture to the training dynamics. Its eigenvalues
are indexed by a frequency, allowing us to obtain a learning rate for each frequency of the
targeted function to be learned, and the ratio between the fastest and the slowest of these rates is a direct
measure of spectral bias in the sense of \citet{rahaman2019spectral}. Whether that ratio
stays bounded as the network capacity grows, and what to do when it does not, is decided by the
symbol alone.

\paragraph{Contributions.}
The contributions of this paper are as follows.
\begin{enumerate}
\item A finite-dimensional, Gram-matrix-based framework for the spectral bias of
linear-in-parameters shallow ReLU models, parameterized by a symmetric bilinear form
$a(\cdot,\cdot)$. Least-squares regression and variational or physics-informed losses are
two instances of one analysis, which ties the \emph{frequency principle}
\citep{xu2020frequency} and the activation-conditioning viewpoint of
\citet{hong2022activation} to classical results on mass and stiffness matrices.

\item A localized architecture, \emph{partition-of-unity gated ReLU}, obtained by
multiplying each ReLU feature by a hat partition of unity. It is still a shallow network,
still linear in its trainable parameters, and its Gram matrix is block-banded
\emph{whatever} bilinear form the loss uses.

\item A spectral characterization by block-Toeplitz analysis, and a bridge from it to the
frequency principle. The theoretical argument is the same for every bilinear form, but the result depends on the
symbol. Reading that symbol at the frequencies the gate grid supports gives a
learning rate per frequency, whose extremes define a \emph{spectral-bias factor}
controlled by the symbol alone. 

\item A preconditioner can be naturally derived from the same symbol with the energy natural-gradient
descent as its dense instance. Its construction shows that what the architecture cannot fix must be left to the optimizer.

\end{enumerate}

\paragraph{Outline.}
Section~\ref{sec:related} surveys spectral-bias and frequency-principle results, NTK-based analyses, architectural remedies, localized linear-in-parameters methods, and the partition-of-unity and block-Toeplitz background on which the framework draws. Section~\ref{sec:problem} sets up the linear-in-parameters model in the abstract bilinear-form setting, recalls how the Gram spectrum drives gradient descent, defines spectral bias formally, and recalls the global ReLU baseline. Section~\ref{sec:gated} introduces the PoU-gated dictionary and establishes its bilinear-form-independent block-banded structure. Section~\ref{sec:spectrum} develops the spectral framework: a scaling lemma, a block-Toeplitz decomposition with a matrix-valued symbol, a frequency-resolved reading of that symbol, and the two specializations to $L^2$ regression (where gating removes the growth of spectral bias with $N$) and to the $H^1$ stiffness matrix (where it does not). Section~\ref{sec:precond} derives the preconditioner needed in the $H^1$. Section~\ref{sec:results} sets up the experiments, fixes the diagnostic and reports the results in both regimes, and Section~\ref{sec:discussion} discusses the implications and the pathways to deeper and higher-dimensional architectures.
\section{Related work}
\label{sec:related}

\paragraph{Spectral bias and the frequency principle.}
The empirical observation that neural networks fit low-frequency content was made first independently by \citet{rahaman2019spectral}, who introduced the term \emph{spectral bias}, and by \citet{xu2019frequency,xu2020frequency}, who phrased it as the \emph{frequency principle} (\textsc{F-principle}). Both lines of work demonstrated, across architectures and datasets, that the relative error in Fourier space decays from low-frequency to high-frequency during training, often spanning several orders of magnitude in time scale. A recent overview is given by \citet{xu2024overview}. The phenomenon is now widely viewed as one of the principal implicit regularization mechanisms of gradient-based deep learning.

\paragraph{NTK-based analyses.}
The dominant theoretical framework for spectral bias is the neural tangent kernel (NTK) of \citet{jacot2018ntk}. In the infinite-width limit, training a network by gradient flow is equivalent to kernel regression with the NTK; consequently, the modes of the target are learned at rates equal to the NTK eigenvalues. \citet{basri2019convergence,basri2020frequency} computed the NTK spectrum on the sphere and on non-uniform data and proved frequency-dependent convergence rates that decay polynomially (uniform data) or worse (non-uniform data). \citet{arora2019fine,du2018gradient,cao2019towards} gave fine-grained convergence and generalization bounds in the same regime. These analyses establish the existence and the severity of spectral bias, but they describe a fixed kernel, and they say little about how a finite-width architecture should be modified to alleviate it. We take a complementary path: we work at finite width, with a fixed feature dictionary, and tie the bias to the symbol of a structured matrix that the architecture controls directly.

\paragraph{Architectural remedies.}
A substantial line of work seeks to mitigate spectral bias by reshaping the input or activation space. Fourier feature mappings \citep{tancik2020fourier} and sinusoidal activations \citep{sitzmann2020implicit} concentrate the mass of the kernel on a tunable frequency band and have become standard for implicit neural representations and PINNs. Kolmogorov--Arnold networks (KANs) and their wavelet variants \citep{liu2024kan,meshir2025wavkan} control the frequency content through learned univariate basis functions, and their spectral behavior has itself been studied \citep{wang2024kanexpressive}. A complementary line of work keeps the architecture fixed and shows how the choice of pointwise nonlinearity shapes the bias: \citet{sahs2025nonlinearity} derive, in the kernel regime, the spectral seminorm that a broad class of activations implicitly penalizes. For PINNs and neural operators specifically, \citet{khodakarami2026spectral} give a systematic empirical study of how architecture, activation, loss design and optimizer interact to produce spectral bias, and extract mitigation guidelines from it; the present paper can be read as supplying, for one architecture class, the closed-form spectral statement behind guidelines of that kind. \citet{wang2021eigenvector, wang2022when} analyze Fourier-feature PINNs through their NTK and propose multi-scale architectures specifically tailored to PDE problems. All of these approaches operate on the kernel/NTK side of the picture; the present work instead targets the \emph{Gram matrix of features} of a shallow ReLU model and exploits classical ideas from approximation theory and conditioning of structured matrices.

\paragraph{Activation- and feature-conditioning viewpoints.}
The closest precursor to the present work is the activation-function study of \citet{hong2022activation}, who isolate the role of the activation by analyzing least-squares problems for shallow networks with global features. They show that the ReLU mass matrix has a condition number scaling like $N^4$ in the number of features and that other activations (e.g.\ ReLU$^p$, smoothed variants) yield more favorable spectra and reduce spectral bias. We use their global-ReLU result as a baseline and extend the conditioning program to a fundamentally different design choice, based on spatial gating, rather than a different point-wise nonlinearity. A related continuum-of-features point of view was developed by \citet{ee2020machine,mei2018meanfield}.

\paragraph{Localized linear-in-parameters methods for PDEs.}
The dictionary studied here is linear in its trainable parameters, and this places it close to a family of methods that solve PDEs by least squares on a fixed, randomly drawn or hand-designed feature set. \citet{dong2021local} combine extreme learning machines with a domain decomposition, representing the solution on each subdomain by a local network whose hidden parameters are frozen and imposing continuity across subdomain interfaces; \citet{chen2022rfm} propose the random feature method, in which random features, a collocation treatment of the PDE and a penalty treatment of the boundary conditions are combined into a single least-squares problem. In both cases the discrete system is explicit, and its conditioning is therefore open to direct analysis, which is the property we exploit here. Our construction differs from these in that the locality is imposed by a partition of unity multiplying each feature, rather than by assigning separate networks to separate subdomains, so that the global dictionary stays conforming and no interface conditions are needed.

\paragraph{Partition of unity, domain decomposition, and block-Toeplitz spectra.}
Partition-of-unity methods \citep{babuvska1997partition} and overlapping Schwarz domain decomposition \citep{Dolean:15:DDM, toselli2005domain} are classical tools in numerical analysis, where they are used to combine local approximation spaces into global ones with controlled overlap. Their spectral signatures are well understood through the theory of block-banded Toeplitz matrices and their matrix-valued symbols \citep{tilli1998note,serra1999spectral,bottcher2005spectral}. Although these tools are ubiquitous in the literature on numerical linear algebra and finite-elements, their use in the analysis of neural networks is much less common. Recent work on finite-basis PINNs \citep{moseley2023fbpinn} introduces overlapping subdomain networks for PDEs, and \citet{heinlein2025ddgaussnewton} observe that such a decomposition makes the otherwise dense Gauss--Newton system block-sparse, but neither provides a spectral analysis of the resulting matrices. Our contribution is to bring the block-Toeplitz machinery into the realm of the Gram matrix of a gated shallow ReLU network and to use it to derive sharp, spectral statements that translate directly into spectral-bias guarantees.

\paragraph{Domain-decomposed and random-feature architectures for SciML.}
The gated dictionary studied here is a linear-in-parameters relative of a family of domain-decomposed SciML architectures developed in earlier work of the author and her collaborators. Multilevel domain-decomposition-based architectures for PINNs \citep{dolean2024multilevel} showed that combining overlapping subdomain networks with coarser levels restores scalability on high-frequency and multi-scale problems, mirroring the classical two-level Schwarz argument. \citet{anderson2026elmfbpinn} replaced the subdomain networks with random features trained by least squares, which makes the resulting discrete systems -- and hence their conditioning -- explicit, and \citet{vanbeek2026filtering} showed that filtering the local feature sets is what keeps such domain-decomposed random-feature methods well conditioned as the decomposition is refined. These works give empirical evidence that localization, combined with an appropriate treatment of the coarse, low-frequency component, is what controls the conditioning of the resulting systems. What they do not provide, and what we aim at here, is a closed-form account of \emph{why} that is so, and of the cases in which localization cannot help.

\section{Problem formulation}
\label{sec:problem}

Let $D=[0,1]$ and let $f^\star\in L^2(D)$ be a target function. We approximate $f^\star$ by a model that is linear in trainable parameters,
\begin{equation}
\label{eq:f-theta}
   f_\theta(x) \;=\; \sum_{\ell=1}^N \theta_\ell\,\varphi_\ell(x),
   \qquad \theta\in\R^N,
\end{equation}
where the dictionary $\{\varphi_\ell\}_{\ell=1}^N\subset L^2(D)$ is fixed (its parameters, e.g.\ ReLU thresholds, are not trained). We work both in the continuous setting, with inner product
$\langle g,h\rangle:=\int_0^1 g(x)\,h(x)\,\dx$, and in an empirical setting with samples $x_1,\dots,x_n\in D$ and $\langle g,h\rangle_n:=\tfrac{1}{n}\sum_{i=1}^n g(x_i)h(x_i)$.

The training loss is the squared $L^2$ error,
\begin{equation}
\label{eq:loss}
  J(\theta) \;=\; \tfrac{1}{2}\,\|f_\theta - f^\star\|^2,
\end{equation}
where $\|\cdot\|$ stands for either the continuous $L^2(D)$ norm or its empirical version. Writing $\sigma(z):=\max\{z,0\}$ for the ReLU activation, our running concrete examples are:
\begin{itemize}
  \item the \emph{global} ReLU dictionary $\varphi^{\mathrm{glob}}_\ell(x)=\sigma(x-s_\ell)$, with knots $0\le s_1<\dots<s_N<1$;
  \item the \emph{gated local} ReLU dictionary introduced in Section~\ref{sec:gated}, where each $\sigma(x-s_\ell)$ is multiplied by a partition-of-unity weight.
\end{itemize}

\begin{remark}
Although \eqref{eq:f-theta} is linear in $\theta$, this is not a degenerate model. With ReLU features it spans a large piecewise-linear function class, and with a fine knot grid it is a universal approximator on $D$. The linear-in-$\theta$ choice is what makes the spectral analysis tractable; it is also exactly the regime captured by the NTK limit of a wide network whose inner weights are frozen at initialization, so the conclusions transfer naturally to the weak-training regime of overparameterized networks.
\end{remark}

\paragraph{Gram matrix and gradient-descent spectral filter.}
Define the Gram matrix $G\in\R^{N\times N}$ and the moment vector $b\in\R^N$ of the dictionary by
\begin{equation}
\label{eq:gram-def}
   G_{\ell\ell'} \;:=\; \langle\varphi_\ell,\varphi_{\ell'}\rangle,
   \qquad
   b_\ell \;:=\; \langle f^\star,\varphi_\ell\rangle.
\end{equation}
A direct expansion of \eqref{eq:loss} gives
\(
   J(\theta) = \tfrac{1}{2}\theta^{\!\top}G\theta - b^{\!\top}\theta + c,
\)
with $c=\tfrac{1}{2}\|f^\star\|^2$. The matrix $G$ is symmetric positive semidefinite, and positive definite whenever the features are linearly independent in the chosen inner product.

Gradient descent with step size $h>0$ on \eqref{eq:loss} reads $\theta^{(\ell+1)}=\theta^{(\ell)}-h\,(G\theta^{(\ell)}-b)$. If $G$ is positive definite, with unique minimizer $\theta^\star=G^{-1}b$, the error $e^{(\ell)}:=\theta^{(\ell)}-\theta^\star$ satisfies the linear iteration
\begin{equation}
\label{eq:error-iter}
   e^{(\ell+1)} \;=\; (I-hG)\,e^{(\ell)}.
\end{equation}

\begin{lemma}[Spectral filter of gradient descent]
\label{lem:gd-filter}
Let $G=U\Lambda U^{\!\top}$ be the eigendecomposition of $G$ with $\Lambda=\mathrm{diag}(\lambda_1,\dots,\lambda_N)$ and $\lambda_1\ge\cdots\ge\lambda_N\ge 0$, and orthonormal eigenvectors $\{u_r\}$. Writing $e^{(\ell)}=\sum_{r=1}^N \alpha_r^{(\ell)} u_r$, the iteration~\eqref{eq:error-iter} yields
\begin{equation}
\label{eq:mode-decay}
   \alpha_r^{(\ell)} \;=\; (1-h\lambda_r)^{\ell}\,\alpha_r^{(0)},
   \qquad r=1,\dots,N.
\end{equation}
If $0<h\le 2/\lambda_1$ and $G$ is positive definite, then $J(\theta^{(\ell)})\to J(\theta^\star)$.

The same statement holds in continuous time. The \emph{gradient flow} of $J$,
\begin{equation}
\label{eq:gradient-flow}
   \dot\theta(t) \;=\; -\nabla J(\theta(t)) \;=\; -\bigl(G\theta(t)-b\bigr),
   \qquad \theta(0)=\theta^{(0)},
\end{equation}
gives $\dot e=-Ge$ for the error, and hence, in the same eigenbasis,
\begin{equation}
\label{eq:mode-decay-flow}
   \alpha_r(t) \;=\; e^{-t\lambda_r}\,\alpha_r(0),
   \qquad r=1,\dots,N.
\end{equation}
\end{lemma}

Both filters --- the discrete $(1-h\lambda_r)^{\ell}$ of \eqref{eq:mode-decay} and the
continuous $e^{-t\lambda_r}$ of \eqref{eq:mode-decay-flow} --- are monotone in
$\lambda_r$ and are governed by the same spectrum. We state results for the flow
\eqref{eq:gradient-flow} from here onwards: the main reason is that no statement about
learning rates is affected by the choice of $h$, and $\ell$ steps of gradient descent
at the largest admissible step $h=1/\lambda_1$ correspond to $t=\ell/\lambda_1$. Every
conclusion transfers to the discrete iteration through \eqref{eq:mode-decay}.

Lemma~\ref{lem:gd-filter} encodes \emph{exactly} the link between the Gram spectrum and training dynamics that we will exploit throughout the paper: each eigenmode of $G$ is a direction in feature space that is learned at a rate $\lambda_r$, so a wide spread of $\{\lambda_r\}$ creates a wide spread of learning timescales. The spectral-bias question becomes a question about $\mathrm{spec}(G)$.

\paragraph{A working definition of spectral bias.}
To make the link between Lemma~\ref{lem:gd-filter} and ``learning low frequencies first'' precise, one needs an appropriate notion of frequency. We use the eigenbasis of a kernel or Gram operator, which generalizes Fourier modes both in the continuum and on finite datasets.
On $\R$, frequency is naturally indexed by the Fourier transform; equivalently, by the eigenfunctions of a smoothing operator. The Gaussian convolution
\[
   (T_s f)(x) := (s/2\pi)^{1/2}\!\int_{\R} e^{-\frac{s}{2}(x-y)^2}f(y)\dy
\]
has Fourier modes $\phi_\omega(x)=e^{i\omega x}$ as eigenfunctions, with eigenvalues $e^{-\omega^2/(2s)}$ that decay rapidly in $|\omega|$. On bounded domains the eigenfunctions of the Laplacian or of a compact integral operator with smooth kernel give an analogous hierarchy, ordering modes from smooth (large eigenvalues) to oscillatory (small eigenvalues).

\paragraph{Data-driven frequencies via Gram matrices.}
On a finite dataset $\mathcal D=\{x_1,\dots,x_n\}$, Fourier modes are not available, but the dictionary itself induces the kernel $k(x,x'):=\sum_{\ell=1}^N\varphi_\ell(x)\varphi_\ell(x')$ whose matrix representation (on samples or in $L^2$) is the Gram matrix $G$. The eigenvectors of $G$ define orthogonal modes of approximation; following \citet{hong2022activation}, we adopt these as the relevant ``frequencies'' of the model.

\begin{definition}[Spectral bias, in our setting]
\label{def:spectral-bias}
Let $\mathcal K$ be a symmetric positive semidefinite operator (a kernel or a Gram matrix) with eigenpairs $\mathcal K\phi_r=\mu_r\phi_r$, $\mu_1\ge\mu_2\ge\dots\ge\mu_N\ge 0$. Expand a target $f^\star=\sum_r a_r^\star \phi_r$ and the model output $f_t=\sum_r a_r(t)\phi_r$ at training time $t$. We say the dynamics exhibits \emph{spectral bias} if $|a_r(t)-a_r^\star|$ decays faster for modes with larger $\mu_r$ than for modes with smaller $\mu_r$.
\end{definition}

In the linear-in-parameters regime, Lemma~\ref{lem:gd-filter} shows that gradient descent realizes \emph{exactly} the spectral filter $(1-h\mu_r)^\ell$ in the Gram eigenbasis, and Definition~\ref{def:spectral-bias} shows that spectral bias is controlled by $\mathrm{spec}(G)$: it
is severe when the $\mu_r$ span many orders of magnitude, and mild when they are
clustered. This is a first and perhaps superficial reading: it measures the spread of the whole spectrum,
whereas what a practitioner observes is the evolution of the frequencies present in the target. Section~\ref{subsec:frequency-law} refines the
definition accordingly.

\paragraph{The global ReLU baseline.}
We briefly recall the conditioning of the standard global ReLU dictionary, which serves as our reference. With uniform knots $s_\ell=(\ell-1)/N$ on $[0,1]$, the continuous Gram matrix
\(
   \Gglob_{\ell\ell'}=\int_0^1\sigma(x-s_\ell)\sigma(x-s_{\ell'})\dx
\)
has explicit entries that are cubic in $\min(s_\ell,s_{\ell'})$ and is closely related to a finite-element mass matrix on a uniform grid \citep{hong2022activation}.

\begin{proposition}[Global ReLU spectrum, after \citet{hong2022activation}]
\label{prop:global-spectrum}
For the uniform-knot global ReLU dictionary on $[0,1]$ with $N$ features, the eigenvalues of $\Gglob$ satisfy
\begin{equation}
\label{eq:global-quartic}
   \frac{\lambda_N(\Gglob)}{\lambda_j(\Gglob)} \;\asymp\; \,\bigl(\frac{N}{j}\bigr)^{4},
   \qquad j=1,\dots,N,
\end{equation}
up to constants independent of $N$. In particular, the condition number satisfies $\kappa(\Gglob)\asymp N^4$.
\end{proposition}

\begin{remark}
The quartic blow-up in \eqref{eq:global-quartic} is the central reason why shallow ReLU networks exhibit strong spectral bias: doubling $N$ shrinks the smallest non-zero eigenvalue by a factor of $\sim 16$, so the slowest-learning mode becomes $16\times$ slower (per gradient-descent iteration) than before. 
\end{remark}

\section{Gated local ReLU dictionaries}
\label{sec:gated}

The architecture we propose retains ReLU activations but replaces the global feature dictionary by a spatially \emph{localized} variant obtained by multiplying each ReLU feature by a partition-of-unity (PoU) weight. The construction is one-dimensional throughout this paper; we discuss tensor-product and higher-dimensional extensions in Section~\ref{sec:discussion}.

\paragraph{Hat partition of unity.}
Fix an integer $K\ge 2$ and define the uniform grid
\(
   x_k:=k/K,\;k=0,\dots,K,
\)
with mesh size $h:=1/K$. The standard hat partition of unity on this grid consists of $K+1$ piecewise-linear functions $\{\psi_k\}_{k=0}^K$. Interior hats ($1\le k\le K-1$) are supported on $[x_{k-1},x_{k+1}]$ and rise linearly from $0$ at the endpoints to $1$ at $x_k$; boundary hats $\psi_0,\psi_K$ are the obvious one-sided versions. By construction
\begin{equation}
\label{eq:pou}
   \psi_k\ge 0,\qquad \sum_{k=0}^K \psi_k(x)=1\quad\text{for all }x\in[0,1].
\end{equation}
Each $\psi_k$ overlaps only with its two nearest neighbors $\psi_{k\pm 1}$, a property that drives the entire spectral analysis.

\paragraph{Gated local ReLU dictionary.}

For each PoU index $k\in\{0,\dots,K\}$ we choose $m$ knots $s_{k,1},\dots,s_{k,m}\in\mathrm{supp}(\psi_k)$ and define the \emph{gated local ReLU} features
\begin{equation}
\label{eq:gated-feature}
   \varphi_{k,j}(x) \;:=\; \psi_k(x)\,\sigma(x-s_{k,j}),
   \qquad k=0,\dots,K,\;\; j=1,\dots,m,
\end{equation}
giving a dictionary of total size $N=(K+1)m$. The associated shallow model is
\begin{equation}
\label{eq:gated-model}
   f^{\mathrm{gate},K}_\theta(x)\;=\;\sum_{k=0}^{K}\sum_{j=1}^{m}\theta_{k,j}\,\varphi_{k,j}(x),
   \qquad \theta\in\R^{(K+1)m}.
\end{equation}
Like the global ReLU model of Section~\ref{sec:problem}, \eqref{eq:gated-model} is linear in the trainable parameters $\theta$; the inner parameters (knots $s_{k,j}$ and gate widths) are fixed by the architecture. The model can be implemented as a shallow network with a hand-crafted first layer that produces gated ReLU features and a linear second layer with weights $\theta$.

\begin{remark}[Gating is not just a different activation]
Replacing $\sigma$ by a different scalar nonlinearity (e.g.\ ReLU$^p$ or hat-shaped activations) changes the smoothness of features but not their support. Gating, in contrast, leaves the activation untouched and changes only the \emph{support} of each feature. As we show next, this is precisely the property that makes the Gram matrix block-banded and amenable to block-Toeplitz analysis. The activation-function and gating viewpoints are therefore complementary: the former reshapes within-feature behavior, the latter reshapes between-feature interactions.
\end{remark}

\paragraph{Block-banded Gram structure.}
\label{subsec:block-banded}

Define the gated Gram matrix
\begin{equation}
\label{eq:gated-gram}
   \Ggate{}^{\!K}_{(k,j),(k',j')} \;:=\; \langle \varphi_{k,j},\varphi_{k',j'}\rangle.
\end{equation}
Because $\mathrm{supp}(\psi_k)\cap\mathrm{supp}(\psi_{k'})=\emptyset$ whenever $|k-k'|>1$, the integrand in \eqref{eq:gated-gram} vanishes identically beyond nearest neighbors. Ordering the unknowns gate by gate (all $j$ for $k=0$, then all $j$ for $k=1$, etc.) yields the following.

\begin{proposition}[Block-banded structure]
\label{prop:block-banded}
The Gram matrix \eqref{eq:gated-gram} has the block-tridiagonal form
\begin{equation}
\label{eq:block-banded-gram}
   \Ggate{}^{\!K} \;=\;
   \begin{pmatrix}
     B_0 & C_0 & & & 0\\
     C_0^{\!\top} & B_1 & C_1 & & \\
     & \ddots & \ddots & \ddots & \\
     & & C_{K-2}^{\!\top} & B_{K-1} & C_{K-1}\\
     0 & & & C_{K-1}^{\!\top} & B_K
   \end{pmatrix},
\end{equation}
where each diagonal block $B_k\in\R^{m\times m}$ encodes within-gate inner products and each off-diagonal block $C_k\in\R^{m\times m}$ encodes the overlap between gates $k$ and $k{+}1$. All other blocks are zero.
\end{proposition}

By contrast, the matrix $\Gglob$ of Section~\ref{sec:problem} is dense: its $(j,k)$ entry is nonzero for every pair of indices. The block-banded form \eqref{eq:block-banded-gram} has only $O((K+1)m^2)$ nonzero entries instead of $O((Km)^2)$.

\paragraph{Connection to finite elements and domain decomposition.}
The structure of \eqref{eq:block-banded-gram} is familiar from numerical analysis. The PoU weights $\{\psi_k\}$ are exactly the standard $P_1$ finite-element basis on the uniform grid, and the gated features \eqref{eq:gated-feature} can be viewed as enrichments of this $P_1$ basis by ReLU functions in the spirit of the partition-of-unity finite-element method \citep{babuvska1997partition}. The block-banded Gram matrix \eqref{eq:block-banded-gram} is, mutatis mutandis, the mass matrix of an enriched $P_1$ space; its spectral conditioning is governed by exactly the tools used in classical numerical analysis. The next section makes this connection quantitative.

\section{Spectral analysis: a unified framework}
\label{sec:spectrum}

The block-banded structure of the gated Gram matrix derived in Section~\ref{sec:gated} is purely a \emph{support} property: it is inherited by any inner product on the dictionary, regardless of whether one weighs interactions by function values or by derivatives. In contrast, the \emph{eigenvalue scaling} depends sharply on which inner product is used. We therefore analyze the gated Gram matrix in a unified way, parameterized by a symmetric bilinear form $a(\cdot,\cdot)$. Two specializations are of central interest:
\begin{itemize}
\item the $L^2$ inner product, which arises in least-squares regression on a target $f^\star$;
\item the $H^1$ semi-inner-product $a(u,v):=\int_0^1 u'(x)v'(x)\,\dx$, which arises in Galerkin discretizations of variational PDEs and in physics-informed losses.
\end{itemize}
The block-Toeplitz argument applies to both. The conclusions do not: gating bounds spectral bias uniformly in $K$ in the regression case, but it does not suffice in the stiffness case, which is what motivates the preconditioners of Section~\ref{sec:precond}.

\subsection{Quadratic losses and the Gram matrix of a bilinear form}
\label{subsec:quadratic-loss}

Let $V\subset L^2([0,1])$ be a function space (typically $L^2$ or $H^1_0$) and $a:V\times V\to\R$ a continuous symmetric bilinear form, positive semidefinite on $V$. Given $\ell\in V'$, the variational problem reads
\begin{equation}
\label{eq:variational-abstract}
   \theta^\star \;=\; \arg\min_{\theta\in\R^N} J^a(\theta),
   \qquad
   J^a(\theta) \;:=\; \tfrac{1}{2}\,a(f_\theta,f_\theta) - \ell(f_\theta)
\end{equation}
where $f_\theta$ is given by \eqref{eq:f-theta} in the global case and \eqref{eq:gated-model} in the gated case.
A direct expansion gives $J^a(\theta)=\tfrac{1}{2}\theta^{\!\top}G^a\theta-b^\top\theta+\text{const}$, with the \emph{Gram matrix of $a$}
\begin{equation}
\label{eq:gram-bilinear}
   G^a_{\ell\ell'} \;:=\; a(\varphi_\ell,\varphi_{\ell'}),
   \qquad b_\ell \;:=\; \ell(\varphi_\ell).
\end{equation}
Gradient descent on \eqref{eq:variational-abstract} obeys exactly the same iteration as before, $e^{(\ell+1)}=(I-hG^a)e^{(\ell)}$, with the more general $G^a$ in place of the $L^2$ Gram matrix. Lemma~\ref{lem:gd-filter} thus generalizes verbatim: each eigenmode of $G^a$ is learned at a rate equal to the corresponding eigenvalue, and the spectral-bias question becomes a question about $\mathrm{spec}(G^a)$.

The two specializations we consider are:

\paragraph{Regression ($a=\langle\cdot,\cdot\rangle_{L^2}$).}
The form $a^{L^2}(u,v):=\int_0^1 uv\dx$ gives $G^{a,L^2}=G^{\mathrm{glob}}$ or $\Ggate{}^{\!K}$ as in Sections \ref{sec:problem} and \ref{sec:gated}; this is the standard mass matrix. The linear functional is $\ell(v)=\langle f^\star,v\rangle_{L^2}$, recovering the $L^2$ regression loss \eqref{eq:loss}.

\paragraph{Variational PDEs ($a$ = energy form).}
For the model 1D Poisson problem $-u''=f$ on $(0,1)$ with homogeneous Dirichlet conditions, the Galerkin formulation seeks $u\in H^1_0$ such that $a^{H^1}(u,v):=\int_0^1 u'v'\dx=\int_0^1 fv\dx$ for all $v\in H^1_0$. Restricting to the gated dictionary gives the \emph{stiffness matrix}
\begin{equation}
\label{eq:stiffness-bilinear}
   A^{\mathrm{gate},K}_{\ell\ell'} \;:=\; \int_0^1 \varphi_\ell'(x)\,\varphi_{\ell'}'(x)\,\dx,
\end{equation}
which is the relevant Gram matrix for the variational formulation. The same matrix appears in physics-informed losses for second-order PDEs, where it controls the Hessian of the energy term~\citep{wang2022when,moseley2023fbpinn}.

\medskip
The block-banded structure of Proposition~\ref{prop:block-banded} holds for \emph{any} bilinear form whose kernel is local enough that integrands vanish outside the union of supports: in particular, both $L^2$ and $H^1$ forms qualify, since each has only the support of $\varphi_\ell\cdot\varphi_{\ell'}$ (resp.\ $\varphi_\ell'\cdot\varphi_{\ell'}'$) as its domain of integration. We can therefore conduct the entire block-Toeplitz analysis once, in the abstract setting, and read off the consequences for each specialization.

\paragraph{Universal $K^{-\alpha}$ scaling lemma}

Under the uniform setup of Section~\ref{sec:gated} (gates $\psi_k(x)=\widehat\psi((x-x_{k-1})/h)$ and knots $s_{k,r}=x_{k-1}+h\tau_r$), a change of variables on each window extracts a common power of $h$ from every entry of the Gram matrix. The exponent depends only on the homogeneity of the integrand, so the argument covers any bilinear form of the form below.

\begin{lemma}[$K^{-\alpha}$ scaling, abstract version]
\label{lem:K-alpha-scaling}
Let $a(u,v)=\int_0^1 \kappa(u,v)\dx$ where $\kappa$ is a (symmetric, bilinear) local pointwise functional homogeneous of degree $\beta$ in scalings $u\mapsto\lambda u$. Assume that $\kappa(\varphi_{k,r},\varphi_{k',s})$, evaluated at gated features \eqref{eq:gated-feature}, has homogeneity $-\delta$ in $h$ relative to $\kappa(\widehat\varphi_r,\widehat\varphi_s)$ on the reference window $[0,1]$. Then there exists a dimensionless symmetric matrix $\widetilde G^{K,m}_a$, independent of $K$, such that
\begin{equation}
\label{eq:K-alpha-scaling}
   G^{a,\mathrm{gate}\,K,m} \;=\; K^{-\alpha}\,\widetilde G^{K,m}_a,
   \qquad \alpha \;=\; 1+\beta-\delta.
\end{equation}
The two cases of interest yield:
\begin{itemize}
  \item $L^2$ form $a(u,v)=\int uv\dx$: each $\varphi_{k,r}=\psi_k\sigma(x-s_{k,r})$ contributes one factor of $h$ from the ReLU ramp under rescaling and one factor of $h$ from $\dx$; total $\alpha=3$.
  \item $H^1$ form $a(u,v)=\int u'v'\dx$: differentiation removes one factor of $h$ from each feature (the chain rule on $\widehat\psi'$ contributes $h^{-1}$, balancing the explicit $h$); arithmetic gives $\alpha=1$.
\end{itemize}
\end{lemma}

\begin{proof}
On window $I_k$, change variables $x=x_{k-1}+h\xi$. For $L^2$, extracts $h^3$. For $H^1$, the derivative of a gated feature is
\(
   \varphi_{k,r}'(x) = h^{-1}\widehat\psi'(\xi)\,(x-s_{k,r})_+ + \widehat\psi(\xi)\,\Theta(x-s_{k,r}),
\)
where $\Theta$ is the Heaviside function. The first term has explicit $h$ from the ReLU, divided by the chain-rule $h^{-1}$, hence $O(1)$ in $h$; the second term is $O(1)$ in $h$ as well. The product $\varphi_{k,r}'\,\varphi_{k',s}'$ is therefore $O(1)$ in $h$, and $\dx=h\,\mathrm d\xi$ contributes the only factor of $h$. Total: $\alpha=1$.\qed
\end{proof}

The exponents $\alpha=3$ ($L^2$) and $\alpha=1$ ($H^1$) are well-known in finite-element analysis and reflect the standard mass-versus-stiffness scaling on a uniform mesh. Here they enter the spectral story directly: they are the rate at which all blocks of the gated Gram matrix shrink as $K\to\infty$.

\subsection{Block-Toeplitz decomposition}
\label{subsec:block-toeplitz-unified}

The interior of the rescaled matrix $\widetilde G^{K,m}_a:=K^{\alpha}G^{a,\mathrm{gate}\,K,m}$ is, by translation invariance of the uniform setup, a block-Toeplitz matrix with constant block-stencil $(C^{\!\top},B,C)$, where $B,C\in\R^{m\times m}$ are determined by the gate profile, the knot pattern and the bilinear form $a$, and are independent of $K$. The boundary blocks differ because $\psi_0,\psi_K$ are one-sided; the resulting correction has rank at most $4m$.

\begin{theorem}[Block-Toeplitz decomposition, abstract form]
\label{thm:block-toeplitz-abstract}
Under the assumptions of Lemma~\ref{lem:K-alpha-scaling}, define $T_{K+1}\in\R^{(K+1)m\times(K+1)m}$ as the block-tridiagonal Toeplitz matrix with stencil $(C^{\!\top},B,C)$. Then
\begin{equation}
\label{eq:rank-perturbation-abstract}
   \widetilde G^{K,m}_a \;=\; T_{K+1} + R_{K,m},
   \qquad \mathrm{rank}(R_{K,m})\le 4m,
   \qquad \|R_{K,m}\|\le M_R,
\end{equation}
with $M_R$ independent of $K$. The Hermitian symbol
\begin{equation}
\label{eq:symbol-abstract}
   f_a(\zeta) \;:=\; C^{\!\top}e^{-i\zeta} + B + Ce^{i\zeta},
\end{equation}
has $m$ real eigenvalue branches $\lambda_1(f_a(\zeta))\le\dots\le\lambda_m(f_a(\zeta))$. With $m_{f_a}:=\min_\zeta\lambda_1$ and $M_{f_a}:=\max_\zeta\lambda_m$,
\begin{equation}
\label{eq:eig-bounds-abstract}
   \mathrm{spec}(T_{K+1})\subset[m_{f_a},M_{f_a}],
   \qquad
   \mathrm{spec}(\widetilde G^{K,m}_a)\subset[m_{f_a}-M_R,\,M_{f_a}+M_R].
\end{equation}
As $K\to\infty$, the eigenvalues of $T_{K+1}$ distribute themselves on the $m$ branches $\zeta\mapsto\lambda_r(f_a(\zeta))$ \citep{ekstrom2018exact,bottcher2005spectral,tilli1998note}.
\end{theorem}

\begin{proof}[Proof sketch]
The block-Toeplitz part is translation-invariant by construction; the boundary correction is supported on the first and last $m$ rows/columns, hence rank $\le 4m$. Locating the spectrum of $T_{K+1}$ inside the symbol range is the classical block-Toeplitz spectral theorem; the band-distribution claim is the corresponding asymptotic statement \citep{ekstrom2018exact}. The Courant--Fischer perturbation by $R_{K,m}$ is uniform in $K$ because $\|R_{K,m}\|$ is bounded independently of $K$. Detailed calculations for $m\le 2$ are given in Appendix~\ref{app:m1-m2}
\end{proof}

Theorem~\ref{thm:block-toeplitz-abstract} applies to any local symmetric bilinear form; the stencils $B$ and $C$, and hence the symbol $f_a$, are the only data that change with the form. We now apply it to the $L^2$ and the $H^1$ case in turn, and the conclusions differ.

\subsection{From the Toeplitz symbol to the frequency principle}
\label{subsec:frequency-law}

The frequency principle of \citet{rahaman2019spectral} and \citet{xu2020frequency} is a
statement about frequencies: the Fourier mode of frequency $q$ is learned at a rate that
decreases with $q$. Theorem~\ref{thm:block-toeplitz-abstract} is a statement about the
spectrum of a block-Toeplitz matrix. The purpose of this subsection is to write down the translation between these two concepts. 

\begin{proposition}[Eigenpairs of the block-Toeplitz part]
\label{prop:toeplitz-eigen}
Let $T_{K+1}$ be as in Theorem~\ref{thm:block-toeplitz-abstract}, with symbol $f_a$,
branches $\lambda_1(f_a(\zeta))\le\dots\le\lambda_m(f_a(\zeta))$, and corresponding symbol
eigenvectors $w_b(\zeta)\in\R^m$. Set
\begin{equation}
\label{eq:zeta-r}
   \zeta_r \;:=\; \frac{r\pi}{K+2},\qquad r=1,\dots,K+1 .
\end{equation}
Then the eigenvectors of $T_{K+1}$ are, blockwise,
$\bigl(\sin(k\zeta_r)\,w_b(\zeta_r)\bigr)_{k=0}^{K}$, with eigenvalues
$\lambda_b(f_a(\zeta_r))$, $b=1,\dots,m$.
\end{proposition}

Together with Lemma~\ref{lem:gd-filter}, this result says that the error along the
eigenvector indexed by $(r,b)$ decays as $e^{-t\lambda_b(f_a(\zeta_r))}$. What it does not
say is which \emph{frequency} that eigenvector represents. Fixing that is a matter of
choice rather than an intrinsic mathematical fact, and the two indices $r$ and $b$ have to
be matched separately.

The index $r$ counts oscillations \emph{across} the gates. Block $k$ of the eigenvector is
supported in $[x_{k-1},x_{k+1}]$ with $x_k=k/K$, so on $[0,1]$ the eigenvector behaves
like $\sin(\pi rx)$ and has $r$ half-waves, whereas $\sin(2\pi qx)$ has $2q$; matching the
counts gives $r=2q$. The index $b$ labels the $m$ branches of the symbol. At a fixed
$\zeta_r$ the $m$ eigenvectors of Proposition~\ref{prop:toeplitz-eigen} all share the same
gate-to-gate oscillation and differ only in the profile $w_b(\zeta_r)\in\R^m$, which is
the shape \emph{within} a single gate across its $m$ knots. Of these $m$ candidate
eigenvectors we keep the one that is closest to the target mode, in the $L^2$ sense made
explicit in \eqref{eq:branch-choice}.

\begin{definition}[Frequency-resolved learning rate]
\label{def:frequency-law}
For $q\lesssim K/2$, set
\begin{equation}
\label{eq:zeta-q}
   \zeta_q \;:=\; \frac{2\pi q}{K+2},
\end{equation}
write $v_{q,b}\in L^2(D)$ for the function represented by the $b$-th eigenvector of
Proposition~\ref{prop:toeplitz-eigen} at $\zeta_q$, expanded in the gated features
$\varphi_{k,j}=\psi_k\,\sigma(\cdot-s_{k,j})$ of \eqref{eq:gated-feature}, and let $b(q)$
be the index of the one closest to the target mode:
\begin{equation}
\label{eq:branch-choice}
   v_{q,b} \;:=\; \sum_{k=0}^{K}\sum_{j=1}^{m}
       \sin(k\zeta_q)\,\bigl(w_b(\zeta_q)\bigr)_j\,\varphi_{k,j},
   \qquad
   b(q) \;:=\; \operatorname*{arg\,max}_{b=1,\dots,m}\;
       \frac{\bigl|\langle v_{q,b},\,\sin(2\pi q\,\cdot)\rangle\bigr|}{\|v_{q,b}\|} .
\end{equation}
The learning rate of frequency $q$, under the gradient flow \eqref{eq:gradient-flow} of
\eqref{eq:variational-abstract}, is then
\begin{equation}
\label{eq:rate-law}
   \boxed{\;R_a(q) \;:=\; \lambda_{b(q)}\bigl(f_a(\zeta_q)\bigr)\;}
\end{equation}
For $m=1$ there is a single branch and $b(q)\equiv1$.
\end{definition}

The restriction $q\lesssim K/2$ is the Nyquist limit of the gate grid: the eigenvector
changes sign at most once per gate, so $\zeta_q\le\pi$. With \eqref{eq:rate-law} in hand the dictionary reads as follows.

\begin{center}
\begin{tabular}{@{}ll@{}}
\toprule
block-Toeplitz object & training object\\
\midrule
symbol variable $\zeta\in[0,\pi]$
   & frequency $q$, through $\zeta_q=2\pi q/(K+2)$\\
branch value $\lambda_b(f_a(\zeta_q))$
   & learning rate $R_a(q)$ of the mode of frequency $q$\\
$f_a$ bounded away from $0$ and $\infty$
   & no frequency ordering (Theorem~\ref{thm:L2-bounds})\\
$f_a(\zeta)\to0$ as $\zeta\to0$
   & the lowest frequency is the slowest (Theorem~\ref{thm:H1-failure})\\
$M_{f_a}/m_{f_a}$
   & the spectral-bias factor (Proposition~\ref{prop:bias-bound})\\
\bottomrule
\end{tabular}
\end{center}
Every row of the table translates a property of the symbol into a property of training. The symbol is fixed
by the architecture and the bilinear form together.

\begin{definition}[Spectral-bias factor]
\label{def:bias-factor}
For a band of frequencies $Q\subseteq\{1,\dots,\lfloor K/2\rfloor\}$, the
\emph{spectral-bias factor} of the bilinear form $a$ on $Q$ is
\begin{equation}
\label{eq:bias-factor}
   \mathcal B_a(Q) \;:=\; \frac{\max_{q\in Q} R_a(q)}{\min_{q\in Q} R_a(q)} ,
\end{equation}
the factor by which the slowest frequency in the band lags the fastest.
$\mathcal B_a(Q)=1$ means no frequency is preferred; spectral bias in the sense of
Definition~\ref{def:spectral-bias} is $\mathcal B_a(Q)\gg1$.
\end{definition}

\begin{proposition}[Symbol bounds control spectral bias]
\label{prop:bias-bound}
Let $m_{f_a}$ and $M_{f_a}$ be as in Theorem~\ref{thm:block-toeplitz-abstract}. For every
band $Q$ and every $K$,
\begin{equation}
\label{eq:bias-chain}
   \mathcal B_a(Q) \;\le\; \kappa(T_{K+1}) \;\le\; \frac{M_{f_a}}{m_{f_a}} ,
\end{equation}
with the convention that the right-hand side is $+\infty$ when $m_{f_a}=0$. There are
accordingly two regimes, separated by the single scalar $m_{f_a}$:
\begin{enumerate}
\item[(i)] \emph{$m_{f_a}>0$.} $\mathcal B_a(Q)$ is bounded by a constant determined by
      the stencil blocks $B$ and $C$ alone, for every band and every $K$. The disparity
      between frequencies does not grow with the size of the model: adding windows adds
      resolvable frequencies without widening it.
\item[(ii)] \emph{$m_{f_a}=0$, attained at $\zeta=0$.} Then $R_a(q)\to0$ as $\zeta_q\to0$,
      so the slowest mode in any band is the \emph{lowest} frequency, and over the full
      resolvable range $Q=\{1,\dots,\lfloor K/2\rfloor\}$ one has
      $\mathcal B_a(Q)=\Theta(K^2)$: the disparity grows quadratically with the size of
      the model.
\end{enumerate}
\end{proposition}

\begin{proof}
Each $R_a(q)$ is by Definition~\ref{def:frequency-law} a value $\lambda_b(f_a(\zeta_q))$,
which by Proposition~\ref{prop:toeplitz-eigen} is an eigenvalue of $T_{K+1}$; a ratio of
two eigenvalues is at most $\kappa(T_{K+1})$, which gives the first inequality. The second
is Theorem~\ref{thm:block-toeplitz-abstract}, which places
$\mathrm{spec}(T_{K+1})\subseteq[m_{f_a},M_{f_a}]$. Part (i) is immediate, since
$m_{f_a}$ and $M_{f_a}$ depend only on $B$ and $C$. For (ii), $f_a$ is smooth and vanishes
at $\zeta=0$, so $R_a(q)\asymp\zeta_q^2\asymp q^2/K^2$ for $q\ll K$ while
$R_a(\lfloor K/2\rfloor)=O(1)$, and the ratio is $\Theta(K^2)$.
\end{proof}

\subsubsection{$L^2$ regression: spectral bias bounded uniformly in $K$}
\label{subsec:L2-application}

Specializing Theorem~\ref{thm:block-toeplitz-abstract} to $a^{L^2}$, for which $\alpha=3$, gives the regression mass matrix. The blocks $B$ and $C$ for $m=1$ and $m=2$ are computed in Appendix~\ref{app:m1-m2}. The property of the symbol $f_{L^2}$ that drives everything in this subsection is that $\min_\zeta\lambda_1(f_{L^2}(\zeta))$ is \emph{strictly positive}.

\begin{theorem}[$L^2$ regression: $K$-uniform eigenvalue and learning-rate bounds]
\label{thm:L2-bounds}
Let $a=a^{L^2}$, under the assumptions of Lemma~\ref{lem:K-alpha-scaling}. There exist
constants $0<c_m^{L^2}\le C_m^{L^2}<\infty$, depending on $m$ but not on $K$, such that
\begin{enumerate}
\item[(a)] \emph{(eigenvalues)} every nonzero eigenvalue $\lambda$ of $\Ggate{}^{\!K}$
      satisfies
      \begin{equation}
      \label{eq:L2-bounds}
         c_m^{L^2}\,K^{-3} \;\le\; \lambda \;\le\; C_m^{L^2}\,K^{-3},
      \end{equation}
      so that $\kappa(\Ggate{}^{\!K})\le C_m^{L^2}/c_m^{L^2}$ uniformly in $K$;
\item[(b)] \emph{(learning rates)} $m_{f_{L^2}}>0$, so
      $R_{L^2}(q)\in[m_{f_{L^2}},M_{f_{L^2}}]$ for every frequency $q$ and every $K$, and
      by Proposition~\ref{prop:bias-bound}
      \begin{equation}
      \label{eq:L2-bias-factor}
         \mathcal B_{L^2}(Q)\;\le\;\frac{M_{f_{L^2}}}{m_{f_{L^2}}}
         \qquad\text{for every band }Q\text{ and every }K .
      \end{equation}
\end{enumerate}
\end{theorem}

Part (b) is the statement that matters for training: the learning rate is bounded above
\emph{and} below independently of frequency, so gating does not merely bound the disparity
of rates, it removes the frequency ordering itself. The constants are explicit for one and
two knots per gate, as seen in the corollaries below.

\begin{corollary}[one knot per gate, $m=1$]
\label{thm:L2-m1}
With one knot per gate ($s_0=0$, $s_k=x_{k-1}$), $\Ggate{}^{\!K}=h^3(T_{K+1}+R_{K+1})$ where $T_{K+1}$ is tridiagonal Toeplitz with diagonal $a_0=11/15$ and off-diagonals $a_1=2/15$, and $\mathrm{rank}(R_{K+1})\le 3$. The eigenvalues of $T_{K+1}$ satisfy
\(
\mu_r(T_{K+1})=a_0+2a_1\cos(r\pi/(K+2))\in[7/15,\,1].
\)
The rescaled mass matrix $\mathcal G_K:=h^{-3}\Ggate{}^{\!K}$ satisfies, for all $K\ge 2$,

\begin{equation}
\label{eq:L2-m1-enclosure}
   \mathrm{spec}(\mathcal G_K)\;\subset\;\Bigl[\tfrac{7-\sqrt{34}}{60},\,1\Bigr]\approx[0.0195,\,1],
   \qquad
   \kappa(\mathcal G_K)\;\le\;4\bigl(7+\sqrt{34}\bigr)\approx 51.3,
\end{equation}
while the learning rates occupy the much smaller interval
\begin{equation}
\label{eq:L2-m1-rates}
   R_{L^2}(q)\;\in\;\Bigl[\tfrac{7}{15},\,1\Bigr]
   \quad\text{for every }q,
   \qquad
   \mathcal B_{L^2}(Q)\;\le\;\tfrac{15}{7}\;\approx\;2.14 .
\end{equation}
\end{corollary}

The two intervals differ because \eqref{eq:L2-m1-enclosure} must also accommodate the
rank-$3$ perturbation $R_{K+1}$ of \eqref{eq:m1-splitting-app}, whose eigenvectors
correspond to no $\zeta_q$: measured, $\kappa(\mathcal G_K)\to33.7$ while
$\mathcal B_{L^2}\le15/7\approx2.14$. The condition number is not an indicator of spectral
bias.

\begin{corollary}[two knots per gate, $m=2$]
\label{thm:L2-m2}
With two knots per gate as in Appendix~\ref{app:m1-m2}, the blocks
\(
B=\begin{psmallmatrix}11/15 & 7/60\\ 7/60 & 1/30\end{psmallmatrix},
\;
C=\begin{psmallmatrix}2/15 & 1/20\\ 0 & 0\end{psmallmatrix}
\)
yield a symbol $f_{L^2}(\theta)$ with $m_{f_{L^2}}\approx 0.005405$ and $M_{f_{L^2}}\approx 1.0279$. With rank-$4$ boundary correction of operator norm $\le 1/20$, every 
eigenvalue of $\mathcal G_K^{(2)}=K^3\Ggate{}^{\!K}$ lies in $[c_2,C_2]$ with $c_2\approx 0.005405$, $C_2\approx 1.028$, and
\begin{equation}
\label{eq:L2-m2-rates}
   R_{L^2}(q)\in[0.005405,\,1.0279],\qquad \mathcal B_{L^2}(Q)\le 190.2 .
\end{equation}
Eigenvalues organize into two branches as $K\to\infty$, with at most four outliers.
\end{corollary}
\begin{remark}[Effect of the gating]
\label{rem:L2-success}
In this case, gating does not equalize the learning rates, but it removes the \emph{growth} of the
disparity with $N$, from $\Theta(N^4)$ for the global dictionary to a bound independent of
$N$. That bound is uniform in $K$ but not in $m$: e.g. we have $33.7$, $190$ and
$9.4\times10^{3}$ at $m=1,2,4$ (Section~\ref{subsec:exp-cond}). Hence the design guideline of
Section~\ref{subsec:design-guidelines}: add windows, not knots.
\end{remark}

\subsubsection{$H^1$ stiffness: spectral bias persists}
\label{subsec:H1-application}

We repeat the construction with $a^{H^1}(u,v)=\int_0^1 u'v'\dx$, obtaining the stiffness
matrix $A^{\mathrm{gate},K}$ of \eqref{eq:stiffness-bilinear}; by
Lemma~\ref{lem:K-alpha-scaling} the universal scaling is now $\alpha=1$. The property of
the symbol $f_{H^1}$ that drives everything here is the opposite of the one in
Section~\ref{subsec:L2-application}: $\min_\zeta\lambda_1(f_{H^1}(\zeta))$ is \emph{zero},
attained at $\zeta=0$. We state this for one knot per gate, where the stencil is explicit.

\begin{theorem}[$H^1$ stiffness: gating resolves neither the conditioning nor the bias]
\label{thm:H1-failure}
Let $a=a^{H^1}$ with one knot per gate, as in Corollary~\ref{thm:L2-m1}. Then:
\begin{enumerate}
\item[(a)] \emph{(stencil and symbol)} the interior entries are
\begin{equation}
\label{eq:H1-m1-stencil}
   A^{\mathrm{gate},K}_{kk} = \tfrac{8}{3}\,h,\qquad
   A^{\mathrm{gate},K}_{k,k+1} = -\tfrac{4}{3}\,h,\qquad
   A^{\mathrm{gate},K}_{kk'}=0\;\;\text{for }|k-k'|>1,
\end{equation}
      with one-sided boundary entries, so the symbol of $h^{-1}A^{\mathrm{gate},K}$ is
      $f_{H^1}(\zeta)=\tfrac{8}{3}(1-\cos\zeta)$ and
\begin{equation}
\label{eq:H1-m1-symbol}
   m_{f_{H^1}} = 0 \;\;(\text{at }\zeta=0),\qquad M_{f_{H^1}} = \tfrac{16}{3} ;
\end{equation}
\item[(b)] \emph{(eigenvalues)} $\mu_r(T_{K+1})=\tfrac{8}{3}(1-\cos\tfrac{r\pi}{K+2})$, so
      $\mu_1\sim\tfrac{4\pi^2}{3}K^{-2}$ while $\mu_{K+1}\to16/3$, and
\begin{equation}
\label{eq:H1-cond}
   \kappa(A^{\mathrm{gate},K}) \;=\; \Theta(K^2) \;=\; \Theta(N^2) ;
\end{equation}
\item[(c)] \emph{(learning rates)}
\begin{equation}
\label{eq:H1-rate-law}
   R_{H^1}(q) \;=\; \tfrac{8}{3}\Bigl(1-\cos\tfrac{2\pi q}{K+2}\Bigr)
   \;\sim\; \frac{16\pi^{2}}{3}\,\frac{q^{2}}{K^{2}}
   \qquad (q\ll K),
\end{equation}
      which \emph{increases} with $q$ and vanishes as $q\to0$, so
      $\mathcal B_{H^1}(\{1,\dots,Q\})\to Q^{2}$ at fixed band and
      $\mathcal B_{H^1}=\Theta(K^2)$ over the full resolvable range.
\end{enumerate}
\end{theorem}

\begin{proof}
The stencil follows by the change of variables of Proposition~\ref{prop:m1-stencil},
applied to $\varphi_k'\varphi_{k'}'$ in place of $\varphi_k\varphi_{k'}$, and the symbol
calculation is direct; (b) and (c) are then Proposition~\ref{prop:toeplitz-eigen} and
Definition~\ref{def:frequency-law} evaluated at $f_{H^1}$.
\end{proof}

Parts (b) and (c) are case (ii) of Proposition~\ref{prop:bias-bound}: the symbol
annihilates the constant eigenvector, so the smallest eigenvalues of $T_{K+1}$ collapse,
and the slowest mode is the \emph{lowest} frequency. Table~\ref{tab:H1-m1-eigs} shows both
growing at the same rate.

\begin{remark}[Effect of the gating]
\label{rem:H1-failure}
In the variational case, gating doesn't change the growth rate: $\kappa$ and $\mathcal B_{H^1}$
are both $\Theta(N^2)$, as for the global dictionary. What it changes is the direction of
the ordering, making $q=1$ the slowest mode. The remedy is therefore not another
dictionary but a preconditioner, which will be built in Section~\ref{sec:precond}. 
\end{remark}

\begin{table}[!htbp]
\centering
\caption{Gated stiffness matrix at $m=1$. The symbol vanishes at $\zeta=0$, so $\lambda_{\min}$
decays as $K^{-2}$ while $\lambda_{\max}$ saturates at $16/3$, and the spectral-bias
factor grows at the same $\Theta(K^2)$ rate: conditioning and frequency disparity degrade
together.}
\label{tab:H1-m1-eigs}
\begin{tabular*}{\tblwidth}{@{}LLLLLR@{}}
\toprule
$K$ & $\lambda_{\min}(h^{-1}A^{\mathrm{gate},K})$ & $\lambda_{\max}(h^{-1}A^{\mathrm{gate},K})$ & $\lambda_{\min}(T_{K+1})$ & $\lambda_{\max}(T_{K+1})$ & $\mathcal B_{H^1}$\\
\midrule
$2^{4}$  & $1.16\!\times\!10^{-2}$ & $5.285$ & $4.05\!\times\!10^{-2}$ & $5.293$ & $32.2$\\
$2^{6}$  & $7.83\!\times\!10^{-4}$ & $5.330$ & $3.02\!\times\!10^{-3}$ & $5.330$ & $441$\\
$2^{8}$  & $4.99\!\times\!10^{-5}$ & $5.333$ & $1.98\!\times\!10^{-4}$ & $5.333$ & $6.74\!\times\!10^{3}$\\
$2^{10}$ & $3.13\!\times\!10^{-6}$ & $5.333$ & $1.25\!\times\!10^{-5}$ & $5.333$ & $1.07\!\times\!10^{5}$\\
$2^{12}$ & $1.96\!\times\!10^{-7}$ & $5.333$ & $7.84\!\times\!10^{-7}$ & $5.333$ & $1.70\!\times\!10^{6}$\\
$2^{14}$ & $1.23\!\times\!10^{-8}$ & $5.333$ & $4.90\!\times\!10^{-8}$ & $5.333$ & $2.72\!\times\!10^{7}$\\
\bottomrule
\end{tabular*}
\end{table}

\label{subsec:diagnosis}
Table~\ref{tab:L2-vs-H1} collects the two cases side by side.

\begin{table}[!htbp]
\centering
\caption{The gated dictionary under the two bilinear forms. The Gram matrices share the
same block-banded, block-Toeplitz structure; only the symbol differs, and its minimum
$m_{f_a}$ decides every row below it. The $\lambda_{\min}$ row agrees by coincidence of
the two scalings, $\Theta(K^{-3})\cdot\Theta(1)$ against $\Theta(K^{-1})\cdot\Theta(K^{-2})$.}
\label{tab:L2-vs-H1}
\begin{tabular*}{\tblwidth}{@{}LLLL@{}}
\toprule
Form $a$ & $L^2$ inner product & $H^1$ semi-inner-product\\
\midrule
Gram matrix     & mass $\Ggate{}^{\!K}$       & stiffness $A^{\mathrm{gate},K}$\\
Universal scale & $K^{-3}$ (Lemma~\ref{lem:K-alpha-scaling}, $\alpha=3$) & $K^{-1}$ ($\alpha=1$)\\
Symbol minimum  & $m_{f_{L^2}}>0$              & $m_{f_{H^1}}=0$ at $\zeta=0$\\
$\lambda_{\min}$ scale & $\Theta(K^{-3})$; uniform after rescaling & $\Theta(K^{-3})$; \emph{not} uniform after rescaling\\
Condition no.\  & $\kappa\le C_m^{L^2}/c_m^{L^2}$, $K$-uniform   & $\kappa(A^{\mathrm{gate},K})=\Theta(N^2)$\\
Spectral-bias factor $\mathcal B_a$ & $\le M_{f_{L^2}}/m_{f_{L^2}}$, $K$-uniform ($2.14$ at $m=1$) & $\Theta(K^2)$; ordering inverted\\
Rate at $q\to0$ & bounded below by $m_{f_{L^2}}>0$ & $R_{H^1}(q)\sim\tfrac{16\pi^2}{3}q^2/K^2\to0$\\
\bottomrule
\end{tabular*}
\end{table}

\section{Preconditioning seen from the optimizer side}
\label{sec:precond}

Proposition~\ref{prop:bias-bound} sets a limit on what the architecture can do: the
spectral-bias factor is bounded by $M_{f_a}/m_{f_a}$, and that bound says nothing once
$m_{f_a}=0$. By Theorem~\ref{thm:H1-failure} this is the situation for $a^{H^1}$, and
changing the gate profile or the knot pattern does not help, because the symbol vanishes
at $\zeta=0$ for a reason that belongs to the bilinear form and not to the dictionary. We
therefore leave the dictionary as it is and change the optimizer instead, replacing the
gradient flow \eqref{eq:gradient-flow} by a preconditioned one.

\paragraph{The preconditioned rate.}
Let $M$ be symmetric positive definite and consider
\begin{equation}
\label{eq:precond-flow}
   \dot\theta \;=\; -M^{-1}\bigl(A^{\mathrm{gate},K}\theta-b\bigr) .
\end{equation}
The error decays in the eigenbasis of $M^{-1}A^{\mathrm{gate},K}$, whose eigenvalues are
those of the symmetric matrix $M^{-1/2}A^{\mathrm{gate},K}M^{-1/2}$. If $M$ is itself
block-Toeplitz with symbol $g$, that matrix has interior symbol
$g^{-1/2}f_ag^{-1/2}$, with the eigenvalues of $g^{-1}f_a$, so
Definition~\ref{def:frequency-law} and Proposition~\ref{prop:bias-bound} apply verbatim to
\begin{equation}
\label{eq:precond-rate}
   R^{M}_a(q) \;:=\; \lambda_{b(q)}\bigl(g(\zeta_q)^{-1}f_a(\zeta_q)\bigr),
   \qquad
   \mathcal B^{M}_a(Q) \;=\;
   \frac{\max_{q\in Q}R^{M}_a(q)}{\min_{q\in Q}R^{M}_a(q)} .
\end{equation}
The architecture fixes $f_a$; the optimizer fixes $g$. The whole of Section~\ref{sec:spectrum}
therefore transfers, with $g^{-1}f_a$ in place of $f_a$, and the design question becomes
the choice of $g$.

The canonical instance is Energy Natural Gradient Descent \citep{muller2023achieving},
which projects the optimal function-space descent direction onto the model's tangent
space. For a linear-in-parameters model with the quadratic loss
\eqref{eq:variational-abstract} this is \eqref{eq:precond-flow} with
$M=A^{\mathrm{gate},K}$ exactly, hence $g=f_a$ and $g^{-1}f_a\equiv I$. The preconditioners
below are computable approximations of that choice.

We consider three approximations to
\[
    P = A^{\mathrm{gate},K} = 
    \begin{pmatrix}
        B_0 & C_0 & & & 0\\
        C_0^{\!\top} & B_1 & C_1 & & \\
        & \ddots & \ddots & \ddots & \\
        & & C_{K-2}^{\!\top} & B_{K-1} & C_{K-1}\\
        0 & & & C_{K-1}^{\!\top} & B_K
   \end{pmatrix},
\]
namely the block diagonal preconditioner 
\[
\diagpre = 
    \begin{pmatrix}
        B_0 &  & & & 0\\
         & B_1 &  & & \\
        &  & \ddots &  & \\
        & &  & B_{K-1} &\\
        0 & & &  & B_K
   \end{pmatrix},
\]
and the block Toeplitz and optimal block circulant (\cite{chan1988optimal}) preconditioners
\[
    \toeppre = 
    \begin{pmatrix}
        B & C & & & 0\\
        C^{\!\top} & B & C & & \\
        & \ddots & \ddots & \ddots & \\
        & & C^{\!\top} & B & C\\
        0 & & & C^{\!\top} & B
   \end{pmatrix},
\qquad
    \circpre = 
    \begin{pmatrix}
        B & \frac{K-1}{K}\,C & & & \frac{K-1}{K}\,C^{\!\top}\\
        \frac{K-1}{K}\,C^{\!\top} & B & \frac{K-1}{K}\,C & & \\
        & \ddots & \ddots & \ddots & \\
        & &\frac{K-1}{K}\, C^{\!\top} & B & \frac{K-1}{K}\,C\\
        \frac{K-1}{K}\,C & & & \frac{K-1}{K}\,C^{\!\top} & B
   \end{pmatrix},
\]
where $B\in\mathbb{R}^{m\times m}$ and $C \in \mathbb{R}^{m\times m}$ are representative
blocks, taken from an interior gate. Each of the three is inserted as $M$ in the
preconditioned flow \eqref{eq:precond-flow}; when the same systems are solved directly, a
preconditioned conjugate gradient iteration is the natural choice, and
Theorem~\ref{prop:precond-exact}(b) below bounds its iteration count.

Of the three, only $\toeppre$ is built from the symbol: its block rows carry the interior
stencil $(B,C)$ of Theorem~\ref{thm:block-toeplitz-abstract}, so its symbol is $g=f_a$,
the choice that \eqref{eq:precond-rate} makes trivial. It agrees with
$A^{\mathrm{gate},K}$ on every interior block, exactly and not approximately, and the two
can differ only on the boundary gates.

\begin{theorem}[block-Toeplitz preconditioning]
\label{prop:precond-exact}
Let $A:=A^{\mathrm{gate},K}$ and let $\toeppre$ be built from the interior stencil
$(B,C)$. Then:
\begin{enumerate}
\item[(a)] \emph{(learning rates)} $g=f_{H^1}$, so $g^{-1}f_{H^1}\equiv I$ and
\begin{equation}
\label{eq:precond-flat}
   R^{\toeppre}_{H^1}(q)\;\equiv\;1,
   \qquad
   \mathcal B^{\toeppre}_{H^1}(Q)\;=\;1
\end{equation}
      for every band $Q$ and every $K$: the frequency dependence is removed exactly, and
      what remains is a boundary effect;
\item[(b)] \emph{(structure)} $A-\toeppre$ is supported on the first and last block rows
      and columns, so
\begin{equation}
\label{eq:precond-rank}
   \mathrm{rank}\bigl(A-\toeppre\bigr)\le 4m,
   \qquad
   \toeppre^{-1}A = I+E,\quad \mathrm{rank}(E)\le 4m .
\end{equation}
      Hence $\toeppre^{-1}A$ has at most $4m+1$ distinct eigenvalues, and in exact
      arithmetic conjugate gradients terminates in at most that many iterations,
      independently of $K$.
\end{enumerate}
\end{theorem}

\begin{remark}[the observed rank]
\label{rem:rank-three}
The rank is often in practice smaller than $4m$ and does not grow with $m$: for the knot pattern of Appendix~\ref{app:m1-m2} we find
$\mathrm{rank}(A-\toeppre)=3$ at every $m\in\{1,2,4,8\}$ and $K\in\{15,31,63\}$ tested.
\end{remark}

\begin{remark}[why only the Toeplitz preconditioner removes the spectral bias]
\label{rem:other-preconditioners}
The symbol of $\diagpre$ is the constant $B$, so
$B^{-1}f_{H^1}$ still vanishes at $\zeta=0$; at $m=1$ it leaves $\mathcal B_{H^1}$
\emph{exactly unchanged} (Table~\ref{tab:precond-cond}). The symbol of $\circpre$ differs
from $f_{H^1}$ at small $\zeta$, which is where $f_{H^1}$ is smallest; it brings
$\mathcal B_{H^1}$ down to a small factor, but one that still grows with $K$. Only
$\toeppre$ removes the frequency dependence, and its condition number is uniform in $K$
when Dirichlet boundary conditions (BC) are used.
\end{remark}

\begin{table}[!htbp]
\centering
\caption{Stiffness matrix with $u(1)=0$ imposed, under the three preconditioners: condition
number on the left, spectral-bias factor over the gate-resolved band on the right. Only
$\toeppre$ is uniform in $K$ in both. It gives $\mathcal B_{H^1}=1$ exactly in every row,
and $\diagpre$ leaves $\mathcal B_{H^1}$ unchanged at $m=1$, so both are omitted on the
right.}
\label{tab:precond-cond}
\begin{tabular*}{\tblwidth}{@{}LL RRRR RR@{}}
\toprule
 &  & \multicolumn{4}{c}{$\kappa$} & \multicolumn{2}{c}{$\mathcal B_{H^1}$}\\
\cmidrule(lr){3-6}\cmidrule(l){7-8}
$m$ & $K$ & unpre. & $\diagpre$ & $\toeppre$ & $\circpre$ & unpre. & $\circpre$\\
\midrule
$1$ & $16$ & $108.4$ & $107.5$ & $13.53$ & $10.68$ & $32.2$ & $2.04$\\
$1$ & $32$ & $423.4$ & $423.1$ & $13.88$ & $11.55$ & $116$ & $2.85$\\
$1$ & $64$ & $1677$ & $1677$ & $14.07$ & $18.96$ & $441$ & $4.47$\\
\midrule
$4$ & $16$ & $425.3$ & $139.4$ & $28.78$ & $22.74$ & $414$ & $3.11$\\
$4$ & $32$ & $852.6$ & $554.4$ & $29.40$ & $24.08$ & $426$ & $4.52$\\
$4$ & $64$ & $3171$ & $2214$ & $29.72$ & $26.34$ & $901$ & $9.32$\\
\midrule
$8$ & $16$ & $4801$ & $145.1$ & $59.27$ & $46.57$ & $4698$ & $14.2$\\
$8$ & $32$ & $4873$ & $559.1$ & $60.59$ & $49.31$ & $4807$ & $12.8$\\
$8$ & $64$ & $7120$ & $2219$ & $61.27$ & $51.94$ & $4899$ & $24.8$\\
\bottomrule
\end{tabular*}
\end{table}

Table~\ref{tab:precond-cond} reports both measures. The $\toeppre$ column of $\kappa$ is
constant in $K$ to three significant figures while $\mathcal B^{\toeppre}_{H^1}=1$.

\begin{remark}[architecture and optimizer are two complementary choices]
\label{rem:arch-vs-opt}
Equation~\eqref{eq:precond-rate} separates the two levers: the dictionary sets the value of $f_a$, the
optimizer sets $g$, and the quantity $g^{-1}f_a$ is now representative of the learning rates. Where $f_a$ is
bounded away from zero, $g=I$ (i.e. no preconditioning) suffices; where it is not, no admissible $f_a$ will help tackling the spectral bias, and the
burden passes to $g$. This is the mechanism behind the finding of
\citet{khodakarami2026spectral} that curvature-aware optimizers reorder the spectral
learning order of physics-informed models while architectural changes alone do not.
\end{remark}

\section{Numerical results}
\label{sec:results}

Two dictionaries are compared at matched parameter count: the global ReLU dictionary
$\varphi_\ell(x)=\sigma(x-s_\ell)$ of Section~\ref{sec:problem}, and the gated dictionary
$\varphi_{k,j}=\psi_k\,\sigma(\cdot-s_{k,j})$ of Section~\ref{sec:gated}, whose double
index we flatten to a single $\ell=1,\dots,N$. In both cases the dictionary is fixed and
only the coefficients $\theta$ of $f_\theta=\sum_\ell\theta_\ell\varphi_\ell$ are trained,
except in Section~\ref{subsec:exp-nonlinear}, where the thresholds are trained as well.

Two problems are solved. For $L^2$ regression, the target is
$f^\star(x)=\sum_{q=1}^{10}A_q\sin(2\pi qx)$ with $A_q=1$, following
\citet[Eq.~8]{rahaman2019spectral}. For the Poisson problem we solve $-u''=f$ on $(0,1)$
with $u(0)=u(1)=0$ and $f=\sum_q(2\pi q)^2A_q\sin(2\pi qx)$, so that the exact solution is
the same sum of sines and the same normalization applies.

\paragraph{The frequency content of the model.}
During training $\theta$ evolves, but what we want to follow is the frequency content of
the function $f_\theta=\sum_\ell\theta_\ell\varphi_\ell$, so we first make that content
explicit. For an integer frequency $q$, set
\begin{equation}
\label{eq:fourier-functionals}
   (s_q)_\ell := 2\!\int_0^1\!\varphi_\ell(x)\sin(2\pi qx)\dx,
   \qquad
   (c_q)_\ell := 2\!\int_0^1\!\varphi_\ell(x)\cos(2\pi qx)\dx ,
\end{equation}
two vectors of $\R^N$ fixed by the dictionary alone. The $q$-th Fourier coefficient of the
model, and the amplitude it carries at that frequency, are
\begin{equation}
\label{eq:amp}
   \tilde f_q(\theta) \;:=\; 2\!\int_0^1\! f_\theta(x)\,e^{-2\pi iqx}\dx
   \;=\; c_q^{\!\top}\theta - i\,s_q^{\!\top}\theta ,
   \qquad
   |\tilde f_q(\theta)| \;=\; \bigl[(s_q^{\!\top}\theta)^2+(c_q^{\!\top}\theta)^2\bigr]^{1/2} .
\end{equation}

\paragraph{How that content evolves.}
With $e(t):=\theta(t)-\theta^\star$ the flow gives $\dot e=-G^ae$, so by
Lemma~\ref{lem:gd-filter} the error decays mode by mode in the eigenbasis
$G^av_r=\mu_rv_r$,
\begin{equation}
\label{eq:flow-eigen}
   e(t) \;=\; \sum_{r=1}^{N} e^{-t\mu_r}\,\langle e(0),v_r\rangle\,v_r .
\end{equation}
Applying  \eqref{eq:fourier-functionals} to \eqref{eq:flow-eigen} gives:
\begin{equation}
\label{eq:coeff-decomp}
   s_q^{\!\top}e(t) \;=\; \sum_{r=1}^{N} e^{-t\mu_r}\,\alpha_{q,r},
   \qquad
   \alpha_{q,r} := (s_q^{\!\top}v_r)\,\langle e(0),v_r\rangle ,
\end{equation}
and likewise for $c_q$. The frequency-$q$ content of the error is therefore governed not
by one rate but by a mixture of all $N$, weighted by $\alpha_{q,r}$ -- that is, by how
much of frequency $q$ each eigenvector $v_r$ carries.

\paragraph{Amplitude rather than error.}
One could follow either the error in the $q$-th Fourier coefficient,
$s_q^{\!\top}e(t)$, or the amplitude $|\tilde f_q(t)|$ that the model has built up there.
The error is of little use here. By \eqref{eq:coeff-decomp} it is eventually
dominated by its slowest term, $s_q^{\!\top}e(t)\sim e^{-t\mu_{\min}}\alpha_{q,\min}$, so
every frequency ends up converging at the same rate $\mu_{\min}$. The
amplitude behaves differently. It rises from $0$ towards $|\tilde f_q(\theta^\star)|$ at a
rate set by the terms with the largest $|\alpha_{q,r}|$. We therefore use the
amplitude, exactly as \citet{rahaman2019spectral} do: color encodes $|\tilde f_q(t)|/A_q$
clipped to $[0,1]$, on a single scale in every figure, with frequency horizontal and
training time vertical as in their Figure~1.

\paragraph{Rise time, and a common clock.}
Each amplitude curve is summarized by the time it takes to reach $90\%$ of its converged
value,
\begin{equation}
\label{eq:risetime}
   t_q \;:=\; \inf\bigl\{\,t\ge0 \;:\; |\tilde f_q(t)| \ge 0.9\,|\tilde f_q(\theta^\star)| \,\bigr\} .
\end{equation}
We measure time in units of the fastest mode,
\begin{equation}
\label{eq:tau}
   \tau \;:=\; \mu_{\max}\,t,\qquad \mu_{\max}:=\lambda_{\max}(G^a),
\end{equation}
which has a concrete reading: $\tau$ counts gradient steps taken at an admissible step
size, so the three panels of a figure are on one clock.

\paragraph{The measured spectral-bias factor.}
Writing $\tau_q=\mu_{\max}t_q$, we report
\begin{equation}
\label{eq:spread}
   \mathcal S(Q) \;:=\; \frac{\max_{q\in Q}\tau_q}{\min_{q\in Q}\tau_q} ,
\end{equation}
which is \eqref{eq:bias-factor} with the predicted rates $R_a(q)$ replaced by the measured
rise times. It is not quite $\mathcal B_a$, and it is not $\kappa$ either; the three
quantities differ only in what they range over.

\begin{center}
\begin{tabular}{@{}lll@{}}
\toprule
 & modes & operator\\
\midrule
$\mathcal B_a(Q)$ & frequency modes only & interior symbol\\
$\mathcal S(Q)$   & frequency modes only & assembled $G^a$\\
$\kappa(G^a)$     & all modes            & assembled $G^a$\\
\bottomrule
\end{tabular}
\end{center}

Both relations follow. First, $\mathcal S\le\kappa(G^a)$ always, because by
\eqref{eq:coeff-decomp} a rise time is bracketed by the fastest and the slowest eigenvalue
that contributes to it. Second, $\mathcal S$ coincides with $\mathcal B_a$ when each
frequency is carried by a single interior eigenvector, since if $\alpha_{q,r}\ne0$ only
for $r=r(q)$ then \eqref{eq:coeff-decomp} gives $t_q=\ln(10)/\mu_{r(q)}$.  Table~\ref{tab:spread-vs-kappa} reports all
three quantities for the configurations used below.

\begin{table}[!htbp]
\centering
\caption{Measured against predicted, at the sizes of Figures~\ref{fig:sb-regression}
and~\ref{fig:sb-pde}. Only $\mathcal S$ comes from the training runs; $\mathcal B_a$ is
read from the symbol over $q=1,\dots,10$, and $\kappa$ from the assembled matrix, neither
needing any training. $\mathcal B_a$ exists only for the gated, block-Toeplitz
dictionaries, and predicts $\mathcal S$ closely where $\kappa$ does not. The $H^1$ rows
are the reduced system the dynamics run on, with the Dirichlet constraint eliminated.}
\label{tab:spread-vs-kappa}
\begin{tabular*}{\tblwidth}{@{}LRRR@{}}
\toprule
configuration & $\mathcal S$ & $\mathcal B_a$ & $\kappa$\\
              & measured      & predicted        & predicted\\
\midrule
$L^2$, global ($N=64$)             & $1.9\times10^{5}$ & ---    & $6.7\times10^{7}$\\
$L^2$, gated ($K=31$, $m=2$)       & $4.5$             & $1.54$ & $189$\\
$H^1$, global ($N=64$)             & $2.23$            & ---    & $824$\\
$H^1$, gated ($K=31$, $m=2$)       & $11.0$            & $10.5$ & $533$\\
$H^1$, gated $+\ \toeppre$         & $4.36$            & $1$    & $14.0$\\
\bottomrule
\end{tabular*}
\end{table}

Sections~\ref{sec:spectrum} and~\ref{sec:precond} already report $\mathcal B_a$ computed
from the symbol, in closed form and without training (Tables~\ref{tab:H1-m1-eigs}
and~\ref{tab:precond-cond}). What follows measures it in the dynamics.

The experiments that follow test different parts of the analysis.
Section~\ref{subsec:exp-cond} tests the eigenvalue statements. Sections~\ref{subsec:exp-regression}
and~\ref{subsec:exp-poisson} test the frequency statements for the two bilinear forms,
where the predictions are opposite: $\mathcal B_{L^2}$ bounded, against
$\mathcal B_{H^1}=\Theta(K^2)$ with the ordering reversed, the second together with its
preconditioned remedy. Section~\ref{subsec:exp-nonlinear} goes beyond the analysis by
training the non-linear components as well, to see whether the conclusion survives once
the dictionary is no longer fixed. Section~\ref{subsec:exp-battery} varies the target,
since $\mathcal S$ is a property of the pair (dictionary, target).

\subsection{Conditioning}
\label{subsec:exp-cond}

We begin with the eigenvalue statements, Theorems~\ref{thm:L2-bounds}(a)
and~\ref{thm:H1-failure}(b), before turning to the frequencies.

\begin{figure}[H]
\centering
\includegraphics[width=0.49\textwidth]{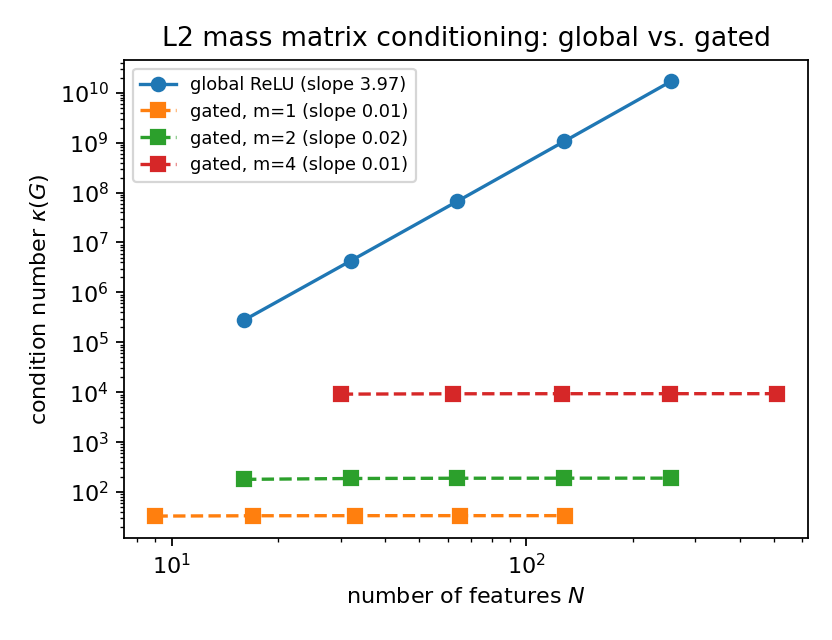}\hfill
\includegraphics[width=0.49\textwidth]{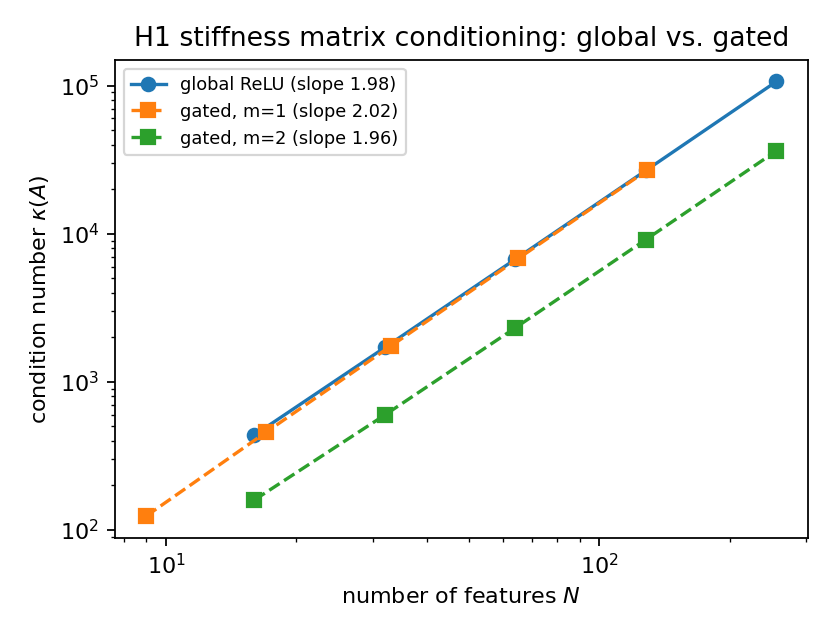}
\caption{Condition number against the number of features $N$. Left, the $L^2$ mass matrix: the
global dictionary grows quartically (slope $3.97$), the gated one stays flat at
$m=1,2,4$. Right, the $H^1$ stiffness matrix: both grow with the same exponent (slopes
$\approx2$). The predictions are Theorems~\ref{thm:L2-bounds}(a)
and~\ref{thm:H1-failure}(b).}
\label{fig:exp-cond}
\end{figure}

The left panel of Figure~\ref{fig:exp-cond} plots $\kappa(G)$ against the number of
features for the $L^2$ mass matrix. A log--log fit gives slope $3.97$ for the global
dictionary, against the $4$ of Proposition~\ref{prop:global-spectrum}, and at most $0.02$
for the gated dictionary at every $m$ tested.

The two curves separate quickly. By $N\approx256$ the global condition number has
reached $1.68\times10^{10}$, while the gated one has barely moved over the whole range:
from $33.09$ to $33.73$ at $m=1$, from $179.9$ to $190.1$ at $m=2$, and from
$9.17\times10^{3}$ to $9.36\times10^{3}$ at $m=4$. The values at the largest size agree
with the limits of Corollaries~\ref{thm:L2-m1} and~\ref{thm:L2-m2}, $\kappa\to33.73$ and
$\kappa\to190.18$, to within $0.03\%$, which checks the analysis and the assembly at
once.

The right panel repeats the study for the stiffness matrix, where the prediction is the
opposite one. Global and gated grow with essentially the same exponent, fitted slopes
$1.98$, $2.02$ and $1.96$, all consistent with the $\Theta(N^2)$ of
Theorem~\ref{thm:H1-failure}. Gating gains a constant factor of about three at fixed $N$,
$2.3\times10^{3}$ against $6.7\times10^{3}$ at $N\approx64$, and nothing beyond that.

\subsection{Regression: gating compresses the frequency-rate spread}
\label{subsec:exp-regression}

\begin{figure}[H]
\centering
\includegraphics[width=0.95\textwidth]{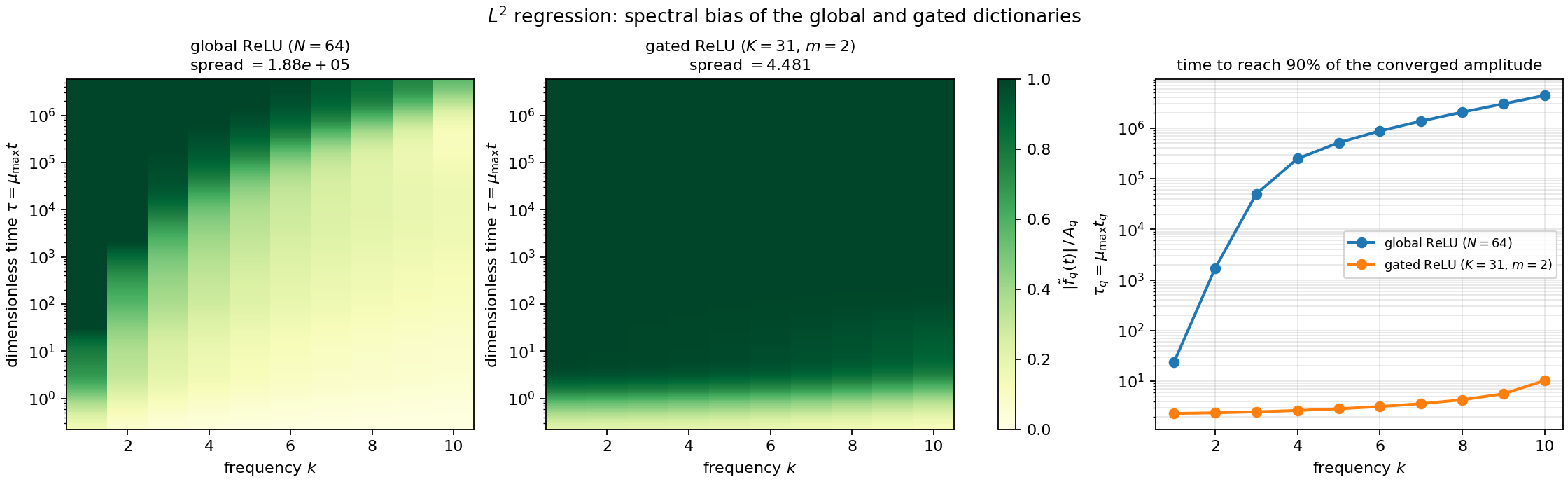}
\caption{$L^2$ regression, linear in parameters. Color is $|\tilde f_q(t)|/A_q$ clipped to
$[0,1]$ against the dimensionless time $\tau=\mu_{\max}t$ of \eqref{eq:tau}, on a log axis
shared across panels. Left: the global dictionary, showing the staircase of
\citet{rahaman2019spectral}, spread $1.9\times10^{5}$. Centre: the gated dictionary, all
ten frequencies together, spread $4.5$. Right: the rise times, whose spread is
$\mathcal S$.}
\label{fig:sb-regression}
\end{figure}

Figure~\ref{fig:sb-regression} shows the amplitudes during training at matched size
($N=64$ global, $K=31$ with $m=2$ gated). The global dictionary reproduces the picture of
\citet{rahaman2019spectral}: a staircase in which each successive frequency darkens an
order of magnitude later than the last, with a spread of $1.9\times10^{5}$ between $q=1$
and $q=10$. For the gated dictionary the same ten frequencies rise together, with a spread
of $4.5$. The right-hand panel shows the same effect in the rise times, which climb through
five decades for the global dictionary and stay inside a factor of five for the gated one.

The symbol predicts $1.54$ for this band, and the remaining factor of three is the
misalignment between the sines and the eigenvectors of $G^a$ discussed above. It is small compared with the four orders of magnitude that separate the two
dictionaries.
What the figure shows is the content of Remark~\ref{rem:L2-success}: gating does not
equalize the rates, it stops their disparity from growing with $N$.

The converged amplitudes call for one caveat. The gated dictionary represents every
frequency of the target to within $0.2\%$, whereas the global one reaches only $0.58$ of
the target amplitude at $q=10$: at $N=64$ it is running out of approximation power at
precisely the frequencies where it is also slowest. Speed and expressivity are therefore
not fully separated in this comparison.

\subsection{Regression with trainable networks}
\label{subsec:exp-nonlinear}

\begin{figure}[H]
\centering
\includegraphics[width=0.95\textwidth]{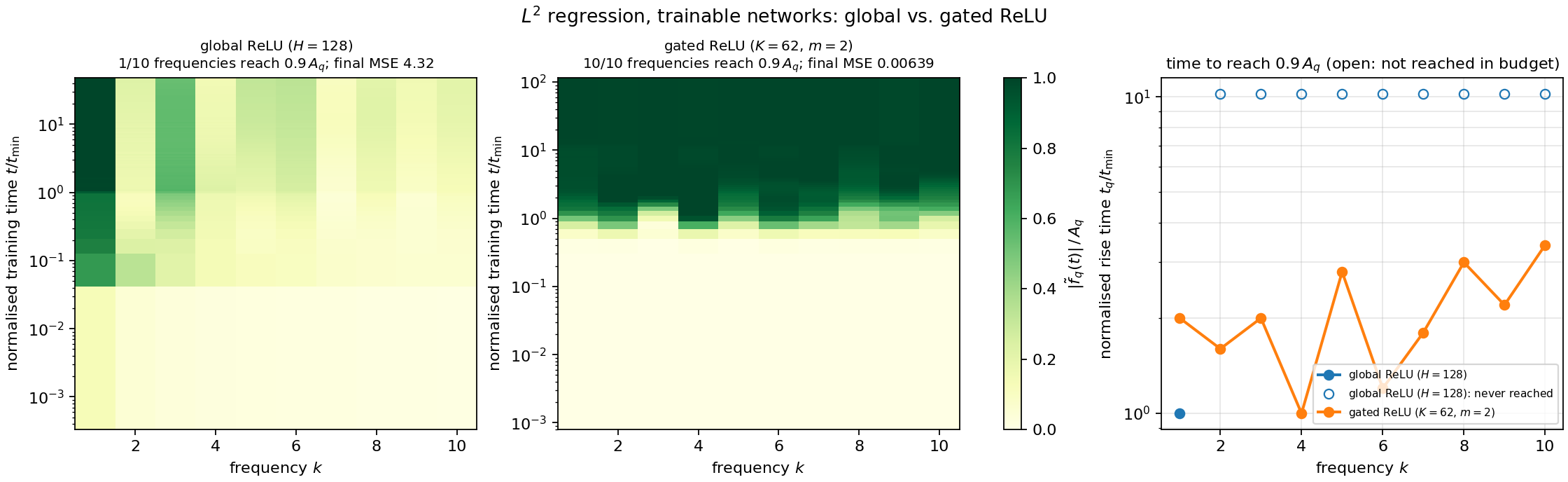}
\caption{$L^2$ regression with trainable networks, same metric and color scale as Figure~\ref{fig:sb-regression}. The gated network resolves all ten frequencies; the global network resolves only $q=1$ within the budget, and open symbols in the right-hand panel mark the frequencies it never reaches.}
\label{fig:sb-nonlinear}
\end{figure}

The experiments above fix the dictionary and train only the output layer.
Figure~\ref{fig:sb-nonlinear} repeats the comparison with every hidden unit's slope,
threshold and output weight trained by full-batch gradient descent, at matched parameter
count, and again with plain gradient descent, since an adaptive optimizer would rescale
the very rates being measured.

The separation is larger than in the linear case. Within the same budget the gated network
brings all ten frequencies to $90\%$ of the target amplitude, taking between
$1.25\times10^{3}$ and $4.25\times10^{3}$ iterations, and reaches a final training MSE of
$6.4\times10^{-3}$. The global network reaches that level at $q=1$ alone; the other nine
frequencies never do, and its final MSE of $4.32$ against a target variance of $5$ means
it has captured about a seventh of the signal.

The mechanism identified by the linear analysis therefore survives in the nonlinear case. In this case, it is clear that The Gram matrix changes along the trajectory and
Section~\ref{sec:spectrum} no longer applies, so this comparison is empirical evidence rather than a verified prediction.

\subsection{Poisson: gating relocates the bias, preconditioning removes it}
\label{subsec:exp-poisson}

\begin{figure}[H]
\centering
\includegraphics[width=0.95\textwidth]{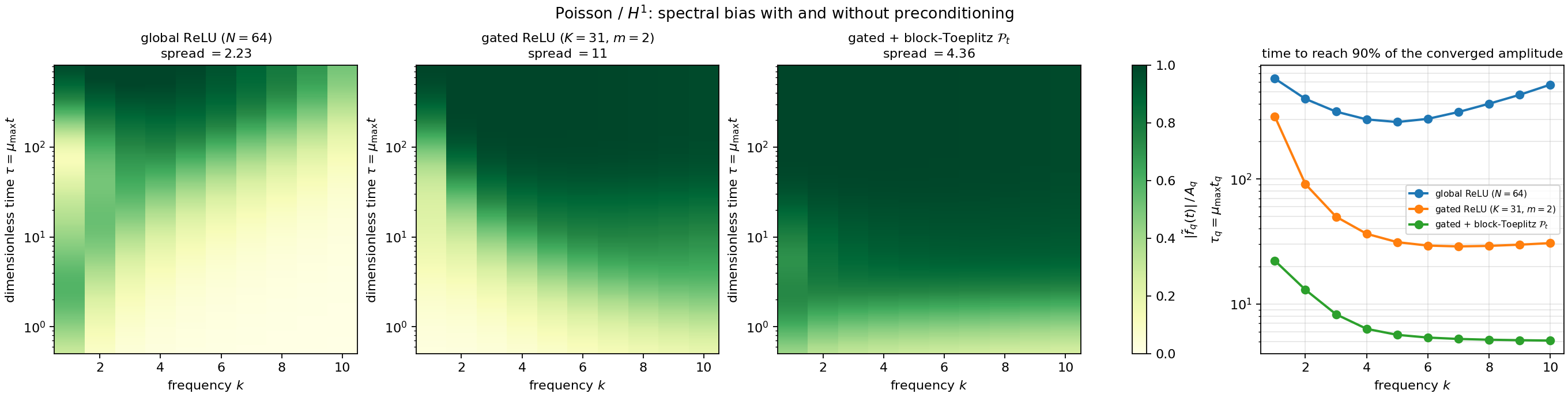}
\caption{Poisson / $H^1$ with $u(0)=u(1)=0$ imposed; same metric and time axis as
Figure~\ref{fig:sb-regression}. Left to right: global (spread $2.23$); gated (spread
$11.0$, ordering inverted, $q=1$ slowest); gated with the block-Toeplitz preconditioner
(spread $4.36$); the rise times. The rising middle panel is $R_{H^1}(q)\propto q^2$ of
Theorem~\ref{thm:H1-failure}(c); the preconditioner flattens it, as
Theorem~\ref{prop:precond-exact}(a) requires.}
\label{fig:sb-pde}
\end{figure}

Figure~\ref{fig:sb-pde} applies the same diagnostic to the Galerkin system, and the
picture is qualitatively different in the way the analysis predicts. For the global
stiffness matrix, the spread is only $2.23$, and the rise-time curve is U-shaped, with both
the lowest and the highest frequencies slow. For the gated matrix, the spread rises to
$11.0$ and the ordering is \emph{inverted}: $q=1$ is now the slowest mode by a factor of
eleven, against the $10.5$ that \eqref{eq:rate-law} predicts for this band, an agreement
to within $5\%$. As a result, in the variational case gating does not remove spectral
bias: it moves the bottleneck to low frequency and makes it more severe.

The block-Toeplitz $\toeppre$ of Section~\ref{sec:precond} cuts the spread to $4.36$,
brings the $q=1$ rise time down by a factor of two and a half and flattens the curve
across the band. Theorem~\ref{prop:precond-exact}(a) predicts
$\mathcal B^{\toeppre}_{H^1}=1$ in the interior, so what remains is the boundary
contribution. In the variational case the remedy is therefore
preconditioning rather than gating, and it follows from the block structure already
established.

\subsection{Robustness to the choice of target}
\label{subsec:exp-battery}

The spread is a property of the pair (dictionary, target), so a single target function is probably not representative for a general behaviour.
Table~\ref{tab:spread-battery} reports it for five targets of increasing difficulty, all
at the sizes of Figure~\ref{fig:sb-regression}: the equal-amplitude sum of
Section~\ref{subsec:exp-regression}; amplitudes increasing with frequency, as in the
second setting of \citet{rahaman2019spectral}; independent random phases, which remove any
alignment between target and grid; a band of high frequencies $k=6,\dots,10$ only; and the
two extremes $k\in\{1,10\}$ alone.

The gated spread never leaves $[2.8,\,6.6]$, while the global one ranges over six orders
of magnitude, from $6.0$ to $5.8\times10^{6}$, and exceeds the gated one in every case.
The margin is widest for a broad band with unequal amplitudes, and narrowest -- a factor
of two -- for the high-frequency band. The last case is instructive: when every
frequency present is high, the global dictionary is uniformly slow rather than
differentially slow, so its \emph{relative} spread is small even though its absolute rates
are not.

\begin{table}[!htbp]
\centering
\caption{Spread $\mathcal S=\max_q\tau_q/\min_q\tau_q$ for $L^2$ regression across five targets, at $N=64$ (global) and $K=31$, $m=2$ (gated). The gated dictionary stays within a factor of seven in every case.}
\label{tab:spread-battery}
\begin{tabular*}{\tblwidth}{@{}LRRR@{}}
\toprule
target & global & gated & ratio\\
\midrule
equal amplitudes                  & $1.9\times10^{5}$ & $4.5$ & $4.2\times10^{4}$\\
increasing amplitudes             & $5.8\times10^{6}$ & $6.6$ & $8.9\times10^{5}$\\
random phases                     & $5.0\times10^{3}$ & $5.9$ & $8.4\times10^{2}$\\
high band only ($k=6,\dots,10$)   & $6.0$             & $2.8$ & $2.1$\\
two extremes ($k\in\{1,10\}$)     & $5.2\times10^{3}$ & $2.9$ & $1.8\times10^{3}$\\
\bottomrule
\end{tabular*}
\end{table}

On the Poisson problem, the benefit of preconditioning is real but target-dependent:
$\toeppre$ cuts the gated spread from $11.0$ to $4.36$ on the equal-amplitude target and
leaves it essentially unchanged, $10.6$ against $10.9$, on the increasing-amplitude one.
Random phases are the revealing case. There the unpreconditioned spread depends strongly
on the draw, ranging from $8.7$ to $67.9$ over five seeds, while the preconditioned one
stays between $1.0$ and $1.4$ for every seed. Built from the interior stencil alone,
$\toeppre$ cancels the vanishing-symbol direction and nothing else a particular right-hand
side may excite, which is exactly the scope of the claim in Section~\ref{sec:precond}.

\section{Discussion and conclusion}
\label{sec:discussion}

One architectural change -- multiplying ReLU features by a partition-of-unity gate --
makes the Gram matrix block-banded for any local bilinear form, and so places a
matrix-valued symbol between the architecture and the training dynamics. That symbol is
what decides the outcome, and the experiments of Section~\ref{sec:results} show it
deciding at the level of the dynamics rather than of matrix spectra alone.

Where the symbol is bounded away from zero, gating alone suffices. For $L^2$ regression it
flattens the condition-number curve against $N$ and compresses the spread of
frequency-dependent rise times from $1.9\times10^{5}$ to $4.5$ at matched size, and the
effect survives when the thresholds are trained as well: the gated network then resolves
all ten frequencies of the target, the global one only the lowest.

Where the symbol vanishes, no amount of gating helps. For the $H^1$ and Poisson case
gating leaves the $\Theta(N^2)$ growth untouched and does more than simply fail: it
\emph{inverts} the ordering, turning $q=1$ into the slowest mode and raising the spread
from $2.23$ to $11.0$. The remedy comes from the same symbol rather than from another
architecture -- a preconditioner read off the interior stencil replaces $f_a$ by
$g^{-1}f_a$ and returns the spread to $4.36$. What the architecture cannot do falls to the
optimizer, and which of the two must act is decided by the bilinear form.

Neither conclusion is tuned to one target: across the five of
Table~\ref{tab:spread-battery} the gated spread stays within $[2.8,6.6]$ while the global
one ranges from $6.0$ to $5.8\times10^{6}$. And the quantity carrying all of this is the
spectral-bias factor, not the condition number; the two coincide only when the extreme
eigenvalues lie at the extreme frequencies.

\subsection{Design guidelines}
\label{subsec:design-guidelines}

The framework of Section~\ref{sec:spectrum} turns the usual trial and error of picking a
shallow architecture into a few explicit recommendations.

\begin{itemize}
\item \textbf{For $L^2$ losses, grow $K$, not $m$.} Once the dictionary is gated, adding
windows adds resolvable frequencies at no conditioning cost, whereas adding knots per
window adds local approximation power without narrowing the gap between frequencies.
Faced with slow high-frequency convergence from a global dictionary, the first thing to
try is a gated one at comparable parameter count, before a different activation, a
different loss weighting, or a larger network.
\item \textbf{For $H^1$ losses, match the preconditioner to the bilinear form.} The
stiffness symbol vanishes at zero frequency whatever the gating, so the $\Theta(N^2)$
conditioning belongs to the second-order form and not to the choice between global and
local features. What restores balanced convergence is a preconditioner built from the
same block stencil that produces the block-Toeplitz structure in the first place
(Section~\ref{sec:precond}).
\item \textbf{Do not use coordinate-adaptive optimizers to test spectral-bias claims.}
Adam's per-coordinate normalization rescales exactly the per-mode rates that
Lemma~\ref{lem:gd-filter} identifies as the carrier of spectral bias, which is what the
adaptive-kernel view of \citet{kopanicakova2026optimization} predicts a coordinate-wise
preconditioner will do to a mode-diagonal problem. We therefore use plain gradient descent
throughout. This concerns how the diagnostic should be read, and is not a claim that Adam should be
avoided in practice.
\end{itemize}

\subsection{Relation to the NTK and to general-purpose optimizers}
\label{subsec:ntk-optimizers}

We analyze plain gradient descent on a fixed dictionary rather than Adam or a
quasi-Newton method, and the reason is worth stating, since much of the SciML optimization
literature is about designing better optimizers for exactly these settings. The survey of
\citet{kopanicakova2026optimization} illustrates this: every first- and second-order
method used for such models is described by the same adaptive kernel
$\Theta_k^{(M_k)}=\tfrac1m J_kM_k^{-1}J_k^\top$, with $M_k=I$ recovering gradient descent,
diagonal second-moment estimates recovering Adam, and the Gauss--Newton or Fisher matrix
recovering natural-gradient methods.

The mode-wise error then decomposes as $e_k=\sum_i c_i(1-\alpha\lambda_i)^kq_i$ in the
eigenbasis of $\Theta_k^{(M_k)}$, which is Lemma~\ref{lem:gd-filter} with $G^a$ in the role
of $\Theta_k^{(I)}$. Every $M_k\neq I$ is therefore an attempt to cluster that spectrum
from the optimizer side rather than the architecture side, and holding $M_k=I$ while
varying the architecture is what isolates the architectural half in closed form. The
preconditioners of Section~\ref{sec:precond} are themselves instances of $M_k\neq I$,
distinguished from general-purpose optimizers only in being read directly off the
block-banded structure of $G^a$ instead of estimated adaptively -- which is what makes them analyzable.

The same view sharpens the role of the two bilinear forms.
\citet{kopanicakova2026optimization} report that for a PINN loss built from a PDE of order
$p$, the conditioning scales as $\kappa\approx(\omega_{\max}/\omega_{\min})^{2p}$. Our
$L^2$ ($p=0$) and $H^1$ ($p=1$) cases are the two smallest instances. Because the
dictionaries here are finite-dimensional and explicit rather than infinite-width limits, we
obtain that law constructively for shallow ReLU-type models, with the symbol responsible
for each exponent written down in Section~\ref{sec:spectrum}.

\subsection{Limitations and pathways forward}
\label{subsec:limitations}

The analysis is deliberately narrow in three respects, each of which points to a next
step rather than to an obstruction.

Everything here is one-dimensional. The block-Toeplitz machinery of
Section~\ref{sec:spectrum} generalizes to block-multilevel-Toeplitz structure in higher
dimensions \citep{serra1999spectral,tilli1998note}, so carrying the partition-of-unity
construction and its symbol to 2D is the natural route toward PDEs on domains.

The boundary treatment of Section~\ref{sec:gated} is not entirely rigorous. A more
accurate variant would sharpen the boundary-layer terms of the decomposition, but it
would not change the interior symbol or anything that follows from it.

We did not extend the trainable-threshold bridge to the $H^1$ case, where training the
thresholds alongside a preconditioned output layer couples fast and slow dynamics in a way
we did not find robust. More broadly, carrying the present closed-form analysis to deep
architectures, where the Gram matrix itself evolves during training, would connect these
results to the NTK evolution literature of Section~\ref{sec:related} and to the
adaptive-kernel framework above. The symbol would no longer be fixed, but it would still
be the object to track.

\FloatBarrier
\section*{Acknowledgements}
The author thanks Jennifer Pestana for helpful discussions on the spectral analysis of
structured matrices and on preconditioning, which shaped the treatment in
Sections~\ref{sec:spectrum} and~\ref{sec:precond}.

\section*{Declaration of competing interest}
The author declares that she has no known competing financial interests or personal
relationships that could have appeared to influence the work reported in this paper.

\section*{Data and code availability}
All matrices in this paper are assembled in closed form from the stencils of
Appendix~\ref{app:m1-m2}, and every experiment uses only \textsc{numpy}, \textsc{scipy}
and \textsc{matplotlib}. The scripts that produce each figure and table, together with the
assembly and diagnostic libraries they depend on, are available from the author on
request.

\appendix

\section{Explicit interior and boundary stencils for $m=1$ and $m=2$}
\label{app:m1-m2}

This appendix contains the explicit computations underlying Corollaries~\ref{thm:L2-m1} and~\ref{thm:L2-m2} and the proof sketch of Theorem~\ref{thm:block-toeplitz-abstract}. Throughout, $h:=1/K$ and $\sigma(z):=\max\{z,0\}$. The integrals below are evaluated on the unit interval $[0,1]$ in the continuous $L^2$ inner product; the empirical version yields the same blocks up to discretization error and is omitted.

\subsection{The case $m=1$}
\label{app:m1}

We choose a single knot per gate, $s_0:=0$ and $s_k:=x_{k-1}$ for $1\le k\le K$, and set $\varphi_k(x):=\psi_k(x)\sigma(x-s_k)$. By Proposition~\ref{prop:block-banded} the Gram matrix $\Ggate{}^{\!K}\in\R^{(K+1)\times(K+1)}$ is tridiagonal up to boundary effects.

\begin{proposition}[Explicit stencil, $m=1$]
\label{prop:m1-stencil}
The entries of $\Ggate{}^{\!K}$ are
\begin{align}
\Ggate{}_{kk}^{\!K}      &= \tfrac{11}{15}\,h^3, & 1\le k\le K-1,\label{eq:m1-diag}\\
\Ggate{}_{k,k+1}^{\!K}   &= \tfrac{2}{15}\,h^3,  & 1\le k\le K-2,\label{eq:m1-off}\\
\Ggate{}_{kk'}^{\!K}     &= 0,                   & |k-k'|>1,\nonumber
\end{align}
together with the boundary entries
\begin{equation}
\label{eq:m1-boundary}
\Ggate{}_{00}^{\!K}=\tfrac{1}{30}\,h^3,
\quad
\Ggate{}_{01}^{\!K}=\tfrac{1}{20}\,h^3,
\quad
\Ggate{}_{KK}^{\!K}=\tfrac{1}{5}\,h^3,
\quad
\Ggate{}_{K-1,K}^{\!K}=\tfrac{2}{15}\,h^3.
\end{equation}
\end{proposition}

\begin{proof}
For $1\le k\le K-1$, the feature $\varphi_k=\psi_k\,\sigma(\cdot-x_{k-1})$ is supported on $I_k\cup I_{k+1}=[x_{k-1},x_{k+1}]$ and is a piecewise quadratic that vanishes at the endpoints. Splitting the integral $\int_0^1\varphi_k^2\dx$ into the contributions from $I_k$ and $I_{k+1}$ and rescaling each to $[0,1]$ via $x=x_{k-1}+h\xi$ reduces every integral to a fixed polynomial integral on $[0,1]$, with three factors of $h$ extracted (two from the ReLU ramp $\sigma(x-x_{k-1})=h\xi$ on the relevant subinterval, one from $\dx$). Summing the two contributions gives $\tfrac{11}{15}\,h^3$, which is~\eqref{eq:m1-diag}. The off-diagonal entry $\Ggate{}_{k,k+1}^{\!K}$ involves only the overlap window $I_{k+1}$ on which $\psi_k$ and $\psi_{k+1}$ are simultaneously nonzero; the same rescaling reduces it to $\tfrac{2}{15}\,h^3$, giving \eqref{eq:m1-off}. The boundary entries~\eqref{eq:m1-boundary} follow from the one-sided supports of $\psi_0$ and $\psi_K$ by the same procedure, with the only difference being that one of the two interior contributions is missing.\qed
\end{proof}

\paragraph{Toeplitz splitting and proof of Corollary~\ref{thm:L2-m1}.}
Define $T_{K+1}\in\R^{(K+1)\times(K+1)}$ as the symmetric tridiagonal Toeplitz matrix with $a_0=11/15$ on the diagonal and $a_1=2/15$ on the sub-/super-diagonal. By Proposition~\ref{prop:m1-stencil},
\begin{equation}
\label{eq:m1-splitting-app}
   \Ggate{}^{\!K} \;=\; h^3\,T_{K+1} + h^3\,R_{K+1},
\end{equation}
where $R_{K+1}$ is supported on rows/columns $\{0,1,K\}$ with explicit entries
\(
(R_{K+1})_{00}=\tfrac{1}{30}-a_0=-\tfrac{7}{10},\;
(R_{K+1})_{01}=(R_{K+1})_{10}=\tfrac{1}{20}-a_1=-\tfrac{1}{12},\;
(R_{K+1})_{KK}=\tfrac{1}{5}-a_0=-\tfrac{8}{15};
\)
all other entries vanish. Hence $\mathrm{rank}(R_{K+1})\le 3$. The eigenvalues of $T_{K+1}$ are explicit \citep{ekstrom2018exact},
\(
\mu_r(T_{K+1})=a_0+2a_1\cos(r\pi/(K+2))\in[a_0-2a_1,a_0+2a_1]=[7/15,1].
\)

Since $\mathrm{rank}(R_{K+1})\le 3$, Weyl's inequality for finite-rank perturbations gives $\lambda_i(\mathcal G_K)\ge\mu_{i+3}(T_{K+1})\ge 7/15$ for $1\le i\le K-2$ (eigenvalues in decreasing order), so at most three eigenvalues of $\mathcal G_K$ fall below $7/15$. For the upper bound, Gershgorin applied to $\mathcal G_K$ gives the disc $[7/15,1]$ for the interior rows, while rows $0$, $1$ and $K$ have centres $1/30$, $11/15$, $1/5$ and radii $1/20$, $11/60$, $2/15$; hence $\lambda_{\max}(\mathcal G_K)\le 1$.

Gershgorin yields no useful lower bound: rows $0$ and $K$ have discs $[-1/60,1/12]$ and $[1/15,1/3]$, which lie \emph{below} $7/15$, the first one not even inside $[0,\infty)$. Nor does the perturbation bound, since $\|R_{K+1}\|_2\approx 0.710>7/15$. The $K$-uniform positive lower bound is obtained instead by a cell-wise coercivity argument. On $I_k$ only $\varphi_{k-1}$ and $\varphi_k$ are supported, and the change of variables $x=x_{k-1}+h\eta$ turns them into the fixed pair $h\,(1-\eta^2,\,\eta^2)$ for $2\le k\le K$, respectively $h\,(\eta(1-\eta),\,\eta^2)$ for $k=1$, with local Gram matrices
\[
   h^3\widehat M=h^3\begin{pmatrix}8/15 & 2/15\\[2pt] 2/15 & 1/5\end{pmatrix},
   \qquad
   h^3\widehat M^{(1)}=h^3\begin{pmatrix}1/30 & 1/20\\[2pt] 1/20 & 1/5\end{pmatrix},
\]
whose smallest eigenvalues are $(11-\sqrt{41})/30\approx 0.1532$ and $(7-\sqrt{34})/60\approx 0.01948$. Since every index occurs in at least one cell,
\[
   v^{\!\top}\mathcal G_K v
   \;=\;\sum_{k=1}^{K}\begin{pmatrix}v_{k-1}\\ v_k\end{pmatrix}^{\!\top}\widehat M^{(k)}\begin{pmatrix}v_{k-1}\\ v_k\end{pmatrix}
   \;\ge\;\min_{k}\lambda_{\min}\bigl(\widehat M^{(k)}\bigr)\sum_{k=1}^{K}\bigl(v_{k-1}^2+v_k^2\bigr)
   \;\ge\;\tfrac{7-\sqrt{34}}{60}\,\|v\|^2 ,
\]
which is the lower bound in~\eqref{eq:L2-m1-enclosure}; the condition-number bound follows from $60/(7-\sqrt{34})=4(7+\sqrt{34})$. The same argument applies verbatim for any $m$ and gives Theorem~\ref{thm:L2-bounds}.\qed

\subsection{The case $m=2$}
\label{app:m2}

For $m=2$ we choose two knots per gate. With $s_{0,1}=0$, $s_{k,1}=x_{k-1}$, $s_{k,2}=x_k$ ($1\le k\le K-1$) and $s_{K,1}=x_{K-1}$, the same change-of-variables computation as in the $m=1$ case gives the explicit interior stencil:

\begin{proposition}[Explicit stencil, $m=2$]
\label{prop:m2-stencil}
For $1\le k\le K-1$, the diagonal block $B=B_k\in\R^{2\times 2}$ and the off-diagonal block $C=C_k\in\R^{2\times 2}$ ($1\le k\le K-2$) of $\Ggate{}^{\!K}$ satisfy
\begin{equation}
\label{eq:m2-blocks-app}
   B \;=\; h^3\begin{pmatrix} 11/15 & 7/60 \\[2pt] 7/60 & 1/30\end{pmatrix},
   \qquad
   C \;=\; h^3\begin{pmatrix} 2/15 & 1/20 \\[2pt] 0 & 0\end{pmatrix}.
\end{equation}
The remaining within-block entries vanish, $\Ggate{}_{(k,1),(k+1,2)}^{\!K}=\Ggate{}_{(k,2),(k+1,2)}^{\!K}=0$. Boundary blocks are
\begin{equation}
\label{eq:m2-boundary}
\begin{gathered}
   \Ggate{}_{(0,1),(0,1)}^{\!K}=\tfrac{1}{30}h^3,\qquad
   \Ggate{}_{(0,1),(1,1)}^{\!K}=\tfrac{1}{20}h^3,\qquad
   \Ggate{}_{(K,1),(K,1)}^{\!K}=\tfrac{1}{5}h^3,\\[2pt]
   \Ggate{}_{(K-1,j),(K,1)}^{\!K}=\Ggate{}_{(k,j),(k+1,1)}^{\!K}\big|_{\text{interior}}.
\end{gathered}
\end{equation}
\end{proposition}

\begin{proof}
Same argument as Proposition~\ref{prop:m1-stencil}, applied separately to each pair of indices $(j,j')\in\{1,2\}^2$. The vanishing entry $C_{12}=0$ reflects the fact that $s_{k+1,2}=x_{k+1}$ falls strictly to the right of $\mathrm{supp}(\psi_k)\cap\mathrm{supp}(\psi_{k+1})=I_{k+1}$, so the corresponding integrand is zero on the overlap. The boundary blocks differ from $B,C$ only because $\psi_0$ and $\psi_K$ are one-sided, and the entries involving $s_{0,2}$ or $s_{K,2}$ that would be nontrivial in the interior are absent at the boundary.
\end{proof}

\paragraph{Block-Toeplitz approximation and proof of Corollary~\ref{thm:L2-m2}.}
Let $T_{K+1}\in\R^{2(K+1)\times 2(K+1)}$ be the block-Toeplitz matrix with constant block-stencil $(C^{\!\top},B,C)$ given by~\eqref{eq:m2-blocks-app} (rescaled by $h^{-3}$, so we work directly with the dimensionless versions). The boundary correction $R_K:=h^{-3}\Ggate{}^{\!K}-T_{K+1}^{\mathrm{cut}}$, where $T_{K+1}^{\mathrm{cut}}$ denotes $T_{K+1}$ with the first and last rows/columns removed, has rank at most $4$ by direct inspection of \eqref{eq:m2-boundary}; its operator norm satisfies $\|R_K\|\le 1/20$.

The matrix-valued symbol is
\(
   f(\theta)=C^{\!\top}e^{-i\theta}+B+Ce^{i\theta},
\)
which for the blocks~\eqref{eq:m2-blocks-app} has two real eigenvalue branches $\theta\mapsto\lambda_1(f(\theta))<\lambda_2(f(\theta))$. Numerical evaluation gives $m_f=\min_\theta\lambda_1\approx 0.005405$ and $M_f=\max_\theta\lambda_2\approx 1.0279$. By the symbol theorem of \citet{ekstrom2018exact,bottcher2005spectral}, every eigenvalue of $T_{K+1}$ lies in $[m_f,M_f]$, and as $K\to\infty$ the eigenvalues distribute themselves on the two branches (approximately uniform sampling of $\theta\in[-\pi,\pi]$ on each branch). Combining with the rank-$4$ Courant--Fischer perturbation $\|R_K\|\le 1/20$ yields the enclosure of Corollary~\ref{thm:L2-m2} with $c_2\ge m_f-\|R_K\|\approx -0.044$, which is non-binding, and $C_2\le M_f+\|R_K\|\approx 1.078$. {The numerically observed values, $c_2\approx 0.005405$ and $C_2\approx 1.028$, are far tighter than these analytic bounds; indeed $c_2$ reproduces $m_f$ to four digits, so the rank-$4$ boundary perturbation does not move the bottom of the spectrum out of the symbol range at all.} \qed

\bibliographystyle{cas-model2-names}
\bibliography{refs}

\end{document}